\documentclass{article}

\PassOptionsToPackage{numbers, sort&compress}{natbib}

\usepackage[preprint]{nips_2018}

\usepackage{graphicx}
\graphicspath{{./Graphics/}}
\usepackage{wrapfig}

\usepackage{subfigure}
\usepackage{tikz,pgf}
\usepackage{epsfig}

\usepackage{setspace}

\usepackage[utf8x]{inputenc} 
\usepackage[T1]{fontenc}   
\usepackage{url}          
\usepackage{booktabs}     
\usepackage{amsfonts}      
\usepackage{amsmath} 
\usepackage{amssymb}
\usepackage{nicefrac} 
\usepackage{microtype}     

\usepackage{tummath}
\usepackage{tumcolors}
\usepackage{xcolor}
\usepackage{siunitx}
\DeclareSIUnit{\px}{px}
\DeclareSIUnit{\GB}{GB}
\DeclareSIUnit{\fps}{fps}
\DeclareSIUnit{\bit}{bit}

\usepackage[pagebackref=true,breaklinks=true,colorlinks,bookmarks=false]{hyperref}

\begin{document}

\title{FutureGAN: Anticipating the Future Frames of Video Sequences using Spatio-Temporal 3d Convolutions in Progressively Growing GANs}

\author{Sandra Aigner and Marco Körner\\ Technical University of Munich\\ Munich, Germany\\ \texttt{\{sandra.aigner, marco.koerner\}@tum.de}}

\maketitle

\begin{abstract}
We introduce a new \emph{encoder-decoder GAN} model, \emph{FutureGAN}, that predicts future frames of a video sequence conditioned on a sequence of past frames.
During training, the networks solely receive the raw pixel values as an input, without relying on additional constraints or dataset specific conditions.
To capture both the spatial and temporal components of a video sequence, spatio-temporal 3d convolutions are used in all encoder and decoder modules. 
Further, we utilize concepts of the existing \emph{progressively growing GAN (PGGAN)} that achieves high-quality results on generating high-resolution single images.  
The FutureGAN model extends this concept to the complex task of video prediction. 
We conducted experiments on three different datasets, \emph{MovingMNIST}, \emph{KTH Action}, and \emph{Cityscapes}. 
Our results show that the model learned representations to transform the information of an input sequence into a plausible future sequence effectively for all three datasets.  
The main advantage of the FutureGAN framework is that it is applicable to various different datasets without additional changes, 
whilst achieving stable results that are competitive to the state-of-the-art in video prediction.
Our code is available at \url{https://github.com/TUM-LMF/FutureGAN}
\end{abstract}

\section{Introduction}
Anticipating a possible future based on experience is an important part of the human decision-making process. 
Simulating this process in machines by teaching them to anticipate future events based on internal representations of the environment could be of relevance for many tasks. 
Automatically predicting the future frames of a video is one approach to tackle this problem.
Video predictions can be of use for planning in robotics, as well as in autonomous driving, especially in reinforcement learning settings. 
They can lead to better decisions, or at least to faster executions, when used as an additional input to the agent. 
As shown by \citet{Mathieu2016}, other tasks, such as object recognition, detection, and tracking, can benefit from the representations that are implicitly learned by such a model.

There are several different approaches that address the pixel-level prediction of video frames. 
Early, often purely deterministic, approaches tended to insufficiently model the uncertainty of the output, which led to blurry predictions. 
Using \emph{generative adversarial networks (GANs)} \citep{Goodfellow2014} is one way to appropriately model the uncertainty of the multi-modal output. 
We build on this idea of training a generative model in an adversarial setting.
GANs learn to model the underlying data distribution implicitly by utilizing a critic, the \emph{discriminator} network, during training time. 
While being trained, the critic constantly provides feedback to the \emph{generator}, whether the generated samples appear real or not. 
This forces the generator to output samples of a similar data distribution as those of the real samples.
\begin{wrapfigure}{r}{0.5\textwidth}
	\centering
	\vspace{5pt}
	\includegraphics[width=0.5\textwidth]{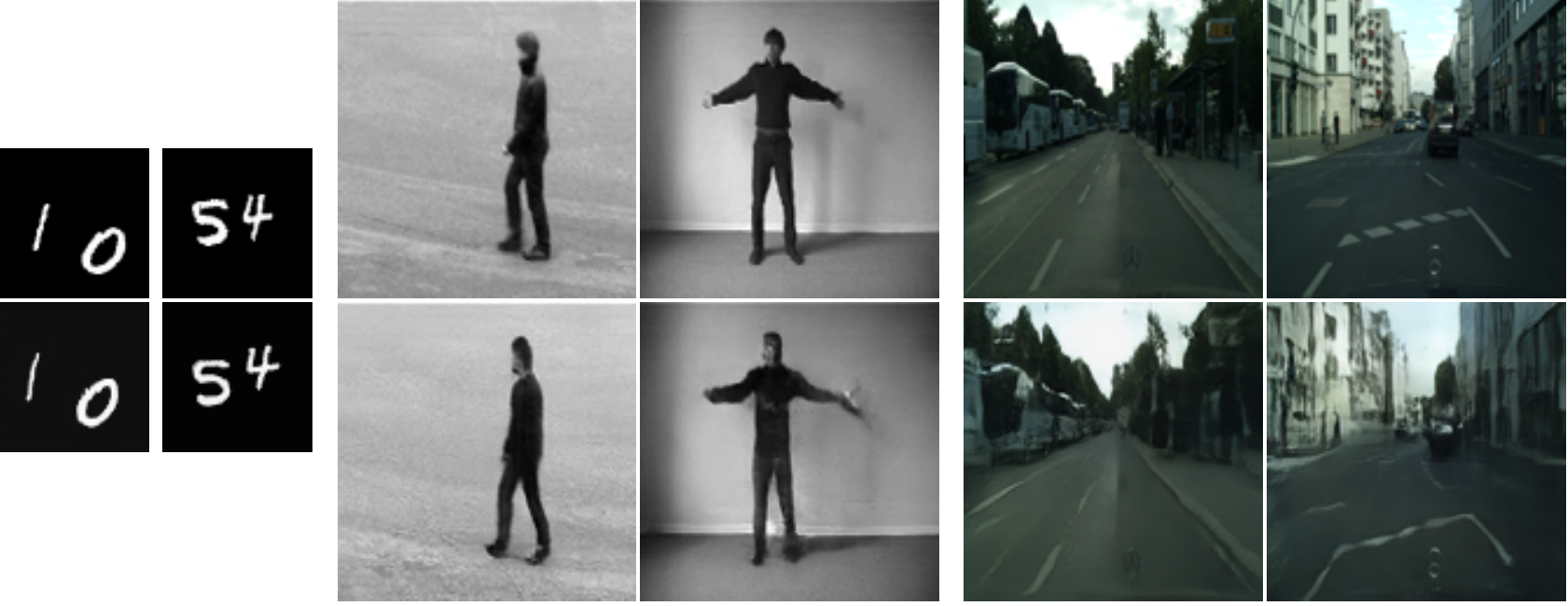}
	\caption{Example predictions for the MovingMNIST, KTH Action, and Cityscapes dataset. Top row: Last observed input frame. Bottom row: Next frames predicted by our FutureGAN.}
	\label{introexample}
	\vspace{-5pt}
\end{wrapfigure}
Although GAN based video prediction methods usually manage to better preserve the sharpness in the predicted frames, there are two major drawbacks. 
First, GANs are hard to train because the training process is highly unstable. 
Secondly, GANs often suffer from mode collapse effects \citep{Salimans2016}, where the generator learns to fool the discriminator by producing samples of a limited set of modes. 
This means, the resulting generative model will not be able to fully capture the underlying data distribution.

In our \emph{FutureGAN} model, we utilize the training strategy of \emph{progressively growing GANs (PGGANs)} \citep{Karras2018} that effectively managed to overcome these problems.
The PGGAN of \citet{Karras2018} was originally designed for generating high-resolution images from a set of random latent variables. 
On this task, it achieved high-quality results. 
The basic principle is to gradually increase the image resolution as the training proceeds by progressively adding layers in both networks. 
For further stabilization of the training, the authors introduced normalization techniques to constrain the signal magnitudes and the competition in both the discriminator and the generator.
We extend this architecture for the complex task of video prediction to benefit from the positive effects on the GAN training.

The primary contribution of this paper is to provide a simple GAN-based model for video prediction that is directly applicable to different datasets using, in general, the same setting.
Our FutureGAN predicts multiple future frames at once when conditioned on a set of past frames.
Contrary to other approaches, both networks solely use the raw pixel value information as an input, without relying on additional conditions. 
To evaluate the FutureGAN framework, we conducted experiments on three datasets of increasing complexity, \textit{i.e.} the \emph{MovingMNIST} dataset \citep{Srivastava2015}, the \emph{KTH Action} dataset \citep{Schuldt2004}, and the \emph{Cityscapes} dataset \citep{Cordts2016}.
Figure \ref{introexample} provides example predictions. 
We show that our model is able to generate plausible futures for all three datasets, while avoiding the problems that typically arise when training GANs.
The predicted frames indicate that the model effectively learned representations of spatial and temporal transformations.

\section{Related Work}
Since 2014, predicting the future frames of a video, from either a single input frame or a sequence of input frames, has become a widely researched topic. 
\citet{Ranzato2014} were the first to provide a baseline model for video prediction with deep neural networks. 
Since then, various other approaches were introduced. 
Most of these combine the raw pixel values of the input frame(s) with learned temporal components \citep{Srivastava2015, Lotter2017, Wang2017, Oliu2018, Liu2017, Vukotic2016, Kalchbrenner2017, Goroshin2015}, dynamically learned filters \citep{DeBrabandere2016}, latent variables \citep{Goroshin2015}, or by explicitly incorporating time dependency \citep{Vukotic2016}.
\citet{Oliu2018}, for example, generate future video frames with a \emph{folded recurrent neural network (fRNN)}. 
This network uses \emph{bijective gated recurrent units (bGRUs)} that learn shared video representations between the encoder and decoder.
Others learn separate representations for the static and dynamic components of a video by adding action or geometry-based conditions, such as pose, optical flow, or depth information \citep{Finn2016, Xue2016, Mahjourian2017, Patraucean2016, Byeon2018, Hao2018, Oh2015}.

The most promising results, especially for long-term predictions, have been achieved just recently by approaches that explicitly include stochasticity in their models \citep{Xue2016, Denton2017, Denton2018, Walker2016, Babaeizadeh2018, Lee2018}. 
Those methods directly address the uncertainty in predicting future frames. 
Generating a set of possible predictions, rather than a single prediction that averages over all modes, prevents from the effect of blurred predictions for an increasing number of time steps.

Another attempt to address the uncertainty in predicting future frames, thus reducing the blurring effect, is to train the generative models in an adversarial setting.
Our approach follows this research branch. 
\citet{Mathieu2016} showed first, networks trained with an adversarial loss term tend to produce sharper results compared to networks only trained on pixel error-based loss metrics, such as the L2 loss.
The idea of using GANs for making video predictions further evolved when traditional image generation GANs were extended for video generation \citep{Vondrick2016b, Saito2017}. 
\citet{Vondrick2016b} use a two-stream network, where foreground and background streams are separated. 
This network generates a sequence of 32 frames using layer-wise spatial and temporal up-sampling with \emph{3d convolutions} \citep{Tran2015}. 
When exchanging the generator's input from random latent variables to the pixel values of an input image, the network then learns to predict future frames.
\citet{Kratzwald2017} build on the approach of \citet{Vondrick2016b}. 
They jointly predict the dynamic and static patterns with an extended Wasserstein GAN (WGAN) \citep{Arjovsky2017}. 
For video generation and prediction, they combine the application-specific L2 loss and the adversarial loss term. 

In contrast to our approach, many GAN-based video prediction methods add additional information, such as motion, content, geometry or action-based conditions, or learn those components separately \citep{Denton2017, Bhattacharjee2017, Tulyakov2018, Chen2017, Xiong2018, Liang2017, Lu2017, Villegas2017, Villegas2017, Vondrick2017, Zeng2017}.
\citet{Villegas2017}, for instance, use an encoder-decoder {convolutional neural network (CNN)} with \emph{convolutional long short-term memory (ConvLSTM)} units to make pixel-level predictions. 
Their network learns to model the motion and content component of the input sequence independently,  using separate encoders.  
GAN approaches often use deterministic \emph{autoencoder (AE)}-based networks with LSTM units where the networks are trained in an adversarial setting.
In many cases, the adversarial term is then added to the loss function \citep{Lotter2016}. 

Mostly related to our approach are \citep{Mathieu2016, Kratzwald2017, Vondrick2016b, Bhattacharjee2017}, but the applied losses and the training strategies differ. 
We follow the idea of using a multi-scale GAN setting for video prediction. 
The idea of a multi-scale or multi-stage GAN for this task has previously been addressed, by either having separate networks, or layer-wise up-sampling operations \citep{Mathieu2016, Bhattacharjee2017, Vondrick2016b, Vondrick2017, Kratzwald2017}.
It is, however, new in this context to add the layers progressively for increasing the image resolution during training.

\section{FutureGAN Model} 

\begin{figure*}[ht]
	\centering
	\includegraphics[width=\linewidth]{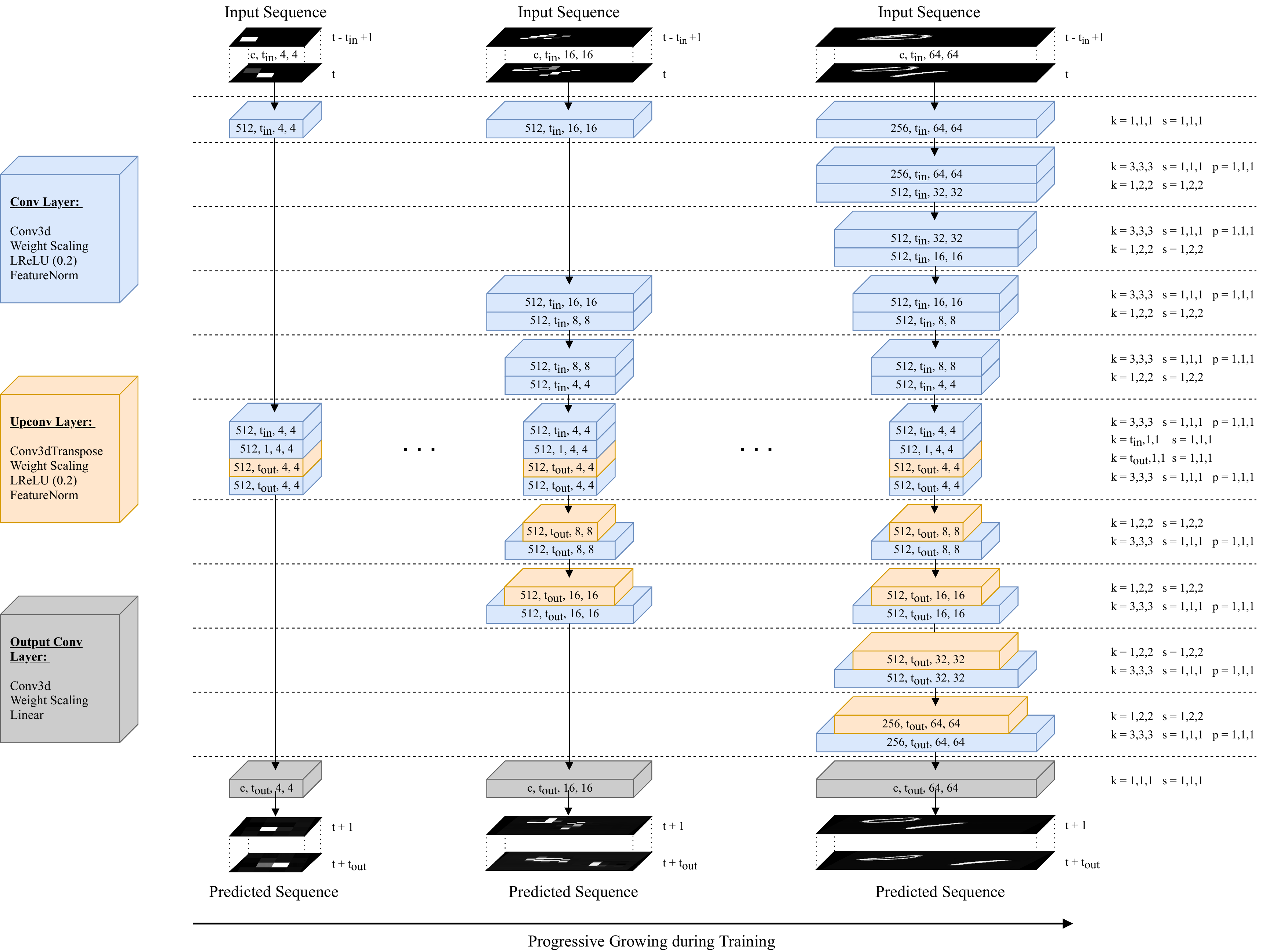}
	\caption{\textbf{FutureGAN generator during training.} 
		We initialize our model to take a set of \SI{4x4}{\px} resolution frames and output frames of the same resolution.
		During training, layers are added progressively to double the resolution after a specified number of iterations. 
		The resolution of the input frames always matches the resolution of the current state of the network.
		This figure illustrates the growth progress of the generator exemplary for the MovingMNIST dataset with a final resolution of \SI{64x64}{\px}. 
		Intermediate \SI{8x8}{\px} and \SI{32x32}{\px} resolution steps are left out for visual clarity.}
	\label{generator}
\end{figure*}

The FutureGAN framework is based on the idea of training a generative model in an adversarial setting and therefore consists of two separate networks. 
Our generator network is trained to predict a sequence of future video frames given a sequence of past frames.
The second network, the discriminator, is trained to distinguish between the generated sequence and a real sequence from the training dataset. 
The discriminator alternately receives real and fake sequences as an input and calculates a score whether the sequence appears real or not.
An output score close to 0 indicates the discriminator rates a given sequence as probably fake.  
The higher the output score of the discriminator for a given sequence, the more realistic it appears to the network. 
The generator network updates its weight parameters according to the feedback it receives from the discriminator, trying to generate sequences that will fool the discriminator. 
Because the training of GANs tended to be highly unstable, we build on the recently proposed PGGAN approach by \citet{Karras2018} that effectively managed to overcome these problems.
We describe the architecture and training strategy of our proposed FutureGAN model in the following.

\subsection{Generator Network}
Our generator network $G$ processes the frames of an input sequence and transforms them into future frames of this sequence.   
The output of the generator can be described as the sequence of future frames 
$\widetilde{\V{x}} = G(\V{z}) = \VecDef{\widetilde{\V{x}}_{t+1}, \ldots , \widetilde{\V{x}}_{t+t_{\text{out}}}}$,
and the input as the sequence of past frames $\V{z}  = \VecDef{\V{x}_{t-t_{\text{in}}+1}, \ldots , \V{x}_{t}}$.
The parameter $t_{\text{in}}$ corresponds to the temporal depth of the input sequence, $t_{\text{out}}$ corresponds to the temporal depth of the output sequence. 
To enable predictions of video frames based on an input sequence, the FutureGAN generator consists of an encoder and a decoder part. 
We extend the PGGAN generator of \citet{Karras2018} to include an encoder that learns a latent representation of the input. 
This latent representation is used by a decoder to generate the predictions. 

For the decoder part of our generator, we modify the basic architecture of the PGGAN generator. 
Figure \ref{generator} illustrates the detailed structure and main components of the FutureGAN generator.
Instead of 2d convolutions, we use 3d convolutions in all convolutional layers. 
This enables the generator to properly encode and decode both the spatial and temporal components of the input sequence. 
Additionally, we realize the spatial upsampling between layers operating on different frame resolutions within a single convolutional layer. 
To perform spatial upsampling only, we use transposed 3d convolutions with asymmetric kernel sizes and strides. 
Originally, \citet{Karras2018} use a nearest neighbor upsampling layer and a convolutional layer separately. 
The encoder part of our generator mirrors the structure of the decoder part, except that the spatial upsampling layers are replaced by spatial downsampling layers. 
We use 3d convolutions with asymmetric kernel sizes and strides to perform spatial downsampling only. 
The bottleneck layers of our generator perform temporal downsampling and upsampling operations to match the temporal depth to the number of input frames and desired output frames, respectively. 
Following the basic design of the PGGAN generator, we add two convolutional layers in the encoder and in the decoder part to increase the network resolution. 
To introduce non-linearity in the networks, \emph{leaky rectified linear units (LReLU)} follow each convolution in the hidden layers.
After each LReLU activation function, a pixel-wise feature vector normalization layer is inserted.

\subsection{Discriminator Network}	
The discriminator of our FutureGAN model is designed to distinguish between real and fake sequences.  
As an input, the discriminator network $D$ alternately receives $\V{x} = \VecDef{\V{x}_{t-t_\text{in}+1}, \ldots , \V{x}_{t+t_{\text{out}}}}$ frames from the training set, representing the ground truth sequence, and 
$\widetilde{\V{x}} = (\V{z}, G(\V{z})) =\VecDef{\V{x}_{t-t_\text{in}+1}, \ldots , \widetilde{\V{x}}_{t+t_{\text{out}}}}$. 
The latter sequence consists of the input and output frames of the generator. 
The output of the discriminator network is a score ${s} = D(\V{x})$ or ${\widetilde{s}} = D(\widetilde{\V{x}})$, respectively. 
This score ranks the given input as either being real or fake. 
We set the labels for the real sequence to $l_{real}=1$ and the labels for fake sequences to $l_{real}=0$. 

Apart from the bottleneck layers, the FutureGAN discriminator closely resembles the encoder part of our generator network. 
One important difference is that there are no pixel-wise feature vector normalization layers in the discriminator. 
Additionally, a mini-batch standard deviation layer is added to one of the last layers. 
\citet{Karras2018} inserted this layer to increase the variation in the generator's outputs, thus to prevent mode collapse.  
This layer computes the standard deviation for each feature in each spatial location over the mini-batch. 
Averaging these values over all features and spatial locations produces a scalar value. 
This value is replicated for every spatial location in the mini-batch, which generates an additional feature map. 
We modify the original layer to calculate this constant feature map for temporal depth as well as spatial locations, in order to increase variation,  especially in the temporal domain. 
To reduce the output of the discriminator to a single scalar, the last layer consists of a fully connected layer, followed by a linear activation function. 
A figure showing the detailed structure and main components of the discriminator is included in appendix \ref{appendix-dsicriminator}.

\subsection{Training Procedure}\label{train}
We initialize our networks to start the training process with a frame resolution of \SI{4x4}{\px}. 
This resolution is gradually increased by a factor of two after the networks have trained for a specified number epochs. 
The number of feature maps in each layer initially is 512. 
Starting from a frame resolution of \SI{64x64}{\px}, the number of feature maps is halved for all newly added layers.
Figure \ref{generator} illustrates the progressive growing for the FutureGAN generator.
Our FutureGAN training closely follows the training procedure described in \citep{Karras2018}.
The next paragraphs briefly introduce the main concepts, for further details we refer to the original paper.

\paragraph{Adding layers for increased resolutions}
Adding new layers to the networks is completed in two steps to ensure a smooth transition between two resolutions. 
The first step is the \emph{transition phase}, where the layers operating on the frames of the next resolution are treated as a residual block whose weights $\alpha$ increase linearly from 0 to 1. 
While the model is in the transition phase, interpolated inputs are fed into both of the networks, making the input frames match the resolution of the current state of the networks. 
The second step is the \emph{stabilization phase}, where the networks are trained for a specified number of iterations before the resolution is doubled again.
Growing the networks progressively both speeds up and stabilizes the training, as the networks only need to learn small transformations between the existing and the newly added layers. 

\paragraph{Weight scaling}
To further stabilize the training, a weight-scaling layer is added on top of all the layers. 
This layer estimates the element-wise standard deviation of the weights and normalizes them to $\widehat{w}_i = w_i / c$, where $w_i$ are the layer weights and $c$ is the normalization constant from \emph{He's initializer} \citep{He2015}. 
Using this layer in a network equalizes the dynamic range, and thus the learning speed, for all weights. 

\paragraph{Feature normalization in the generator}
Another element for stabilizing the training process is the pixel-wise feature vector normalization in the generator.
This element follows the activations of each convolutional layer. 
Based on a variant of the \emph{local response normalization} \citep{Krizhevsky2012}, the feature vector is normalized to unit length in each pixel. 
To make this layer applicable to the FutureGAN generator, we modified it to operate on both the spatial and temporal elements of the feature maps.  
The procedure can be described as $\V{b}_{x,y,z}=\V{a}_{x,y,z} / \sqrt{\frac{1}{n_{\text{f}}} {\V{a}_{x,y,z}}^\top \V{a}_{x,y,z}+  \epsilon}$, where $\epsilon=10^{-8}$, $n_{\text{f}}$ is the number of feature maps, $\V{a}_{x,y,z}$ is the original, and $\V{b}_{x,y,z}$ the normalized feature vector of the pixel $(x,y,z)$. 
Using this technique prevents the escalation of signal magnitudes in the generator and discriminator that result from an unhealthy competition between the two networks.

\paragraph{WGAN-GP loss with epsilon penalty}
Our loss function consists of the \emph{Wasserstein GAN with gradient penalty (WGAN-GP)} loss \citep{Gulrajani2017} and an additional term to prevent the loss from drifting, the epsilon-penalty term. 
Using the WGAN-GP loss effectively increases the quality of the generated frames.

The WGAN-GP loss with epsilon penalty for optimizing the discriminator is defined as  
\begin{equation}
	\begin{aligned}
	L_D(\V{x}, \widetilde{\V{x}}, \widehat{\V{x}}) = \underbrace{\underset{\widetilde{\V{x}} \sim \mathbb{P}_g}{\mathbb{E}}[D(\widetilde{\V{x}})] - \underset{\V{x} \sim \mathbb{P}_r}{\mathbb{E}}[D(\V{x})]}_{\text{WGAN loss}} + \underbrace{\lambda \underset{\widehat{\V{x}} \sim \mathbb{P}_{\widehat{\V{x}}}}{\mathbb{E}}[(\|{\nabla_{\widehat{\V{x}}} D(\widehat{\V{x}})}\|_2-1)^2]}_{\text{gradient-penalty}} + \underbrace{\varepsilon \underset{\V{x} \sim \mathbb{P}_r}{\mathbb{E}}D(\V{x})^2}_{\text{epsilon-penalty}}, 
	\end{aligned}
\end{equation}
where $\mathbb{P}_r$ is the data distribution, $\mathbb{P}_g$ is  the  model  distribution  implicitly  defined  by $\widetilde{\V{x}} = G(\V{z})$, $\widetilde{\V{x}} \sim p(\widetilde{\V{x}})$, $\varepsilon$ is the epsilon-penalty coefficient, and $\lambda$ is the gradient-penalty coefficient.
$\mathbb{P}_{\widehat{\V{x}}}$ is implicitly defined, sampling uniformly along straight lines between pairs of points sampled from the data distribution $\mathbb{P}_r$ and the generator distribution $\mathbb{P}_g$. 

The WGAN(-GP) loss for optimizing the generator is defined as 
\begin{align}
L_G(\widetilde{\V{x}}) = - \underset{\widetilde{\V{x}} \sim \mathbb{P}_g}{\mathbb{E}}[D(\widetilde{\V{x}})].
\end{align}

\section{Experiments and Evaluation}
We conducted experiments on three datasets of increasing complexity, \textit{i.e.} the MovingMNIST dataset \citep{Srivastava2015}, the KTH Action dataset \citep{Schuldt2004}, and the Cityscapes dataset \citep{Cordts2016}.
The experiments on the MovingMNIST and the KTH Action dataset were carried out on an NVIDIA Tesla P100 GPU with \SI{16}{\GB} of RAM.
For the experiments on the Cityscapes dataset, we used an NVIDIA Titan X Pascal GPU with \SI{12}{\GB} RAM.
The FutureGAN model is implemented in PyTorch. 
For the optimization, we used the \emph{ADAM optimizer} \citep{Kingma2015} with $\beta_1 = 0.0$ and $\beta_2 = 0.99$. 
Our initial learning rate was heuristically set to $l = 0.001$.
Every resolution step, we adjusted the batchsize dynamically during training, according to available GPU RAM.
Therefore, we decay our learning rate by a factor of $0.87$ in each resolution step. 
The penalty coefficients of the WGAN-GP loss with epsilon-penalty were set to $\lambda=10$ and $\varepsilon=0.001$, as proposed in \citep{Karras2018}.  

On the MovingMNIST dataset, we trained our network until a resolution of \SI{64x64}{\px}. 
The resolution of the MovingMNIST data already matched the final network resolution. 
For the KTH Action dataset, and the Cityscapes dataset, we used a final resolution of \SI{128x128}{\px}. 
The original size of the KTH Action videos is \SI{160x120}{\px}, the Cityscapes frames have an original size of \SI{2048x1024}{\px}. 
These resolutions did not match the size of our final network resolution.
Therefore, we resized all frames of both datasets bicubically to a resolution of \SI{128x128}{\px}, beforehand.  
During training, the frames were downsampled to match the current resolution of the networks using nearest neighbor interpolation. 
All networks were trained for 10 epochs each in transition and stabilization phase of every resolution step, and another 20 epochs in the final phase. 
This results in a total number of training epochs of 120 for MovingMNIST, and 140 for KTH Action and Cityscapes. 

To evaluate the networks quantitatively, we provide values for the \emph{mean squared error (MSE)}, the \emph{peak signal-to-noise ratio (PSNR)}, and \emph{structural similarity index (SSIM)} between the ground truth and the predicted frame sequence. 
We compare our FutureGAN models to a naive baseline of simply copying the last frame of the input sequence, as well as to state-of-the-art approaches. 
Additionally, we provide comparisons between the optical flow maps of the ground truth and the predicted frames in appendix \ref{appendix-opticalflow}. 

\begin{table}[h]
	\caption{Average results over 6 frames for the full test splits (best results are bold)}
	\label{movingmnist_kthaction_avg}
	\centering
	\small
	\begin{tabular}{lrrrrrr}
		\cmidrule(r){1-5}
		& \multicolumn{2}{c}{MovingMNIST} & \multicolumn{3}{c}{KTH Action} \\
		\cmidrule(r){2-5}
		& \multicolumn{1}{c}{MSE} & \multicolumn{1}{c}{SSIM} & \multicolumn{1}{c}{MSE} & \multicolumn{1}{c}{SSIM} \\
		\cmidrule(r){1-5}	
		CopyLast 					& 0.2580  			& 0.6791 				& 0.0116		& 0.8532 \\
		FutureGAN (ours)   			& \textbf{0.1603} 	& \textbf{0.7780}   	& 0.0120		& 0.6180 \\
		fRNN \citep{Oliu2018}		& 0.1854 			& 0.7408    			& 0.0181		& 0.7053 \\
		MCNet \citep{Villegas2017}	&     -- 			&     --    	& \textbf{0.0048}		& \textbf{0.8692} \\
		\cmidrule(r){1-5}
	\end{tabular}
\end{table}

\begin{figure*}[h]
	\centering
	\subfigure{
		\includegraphics[width=0.3\linewidth]{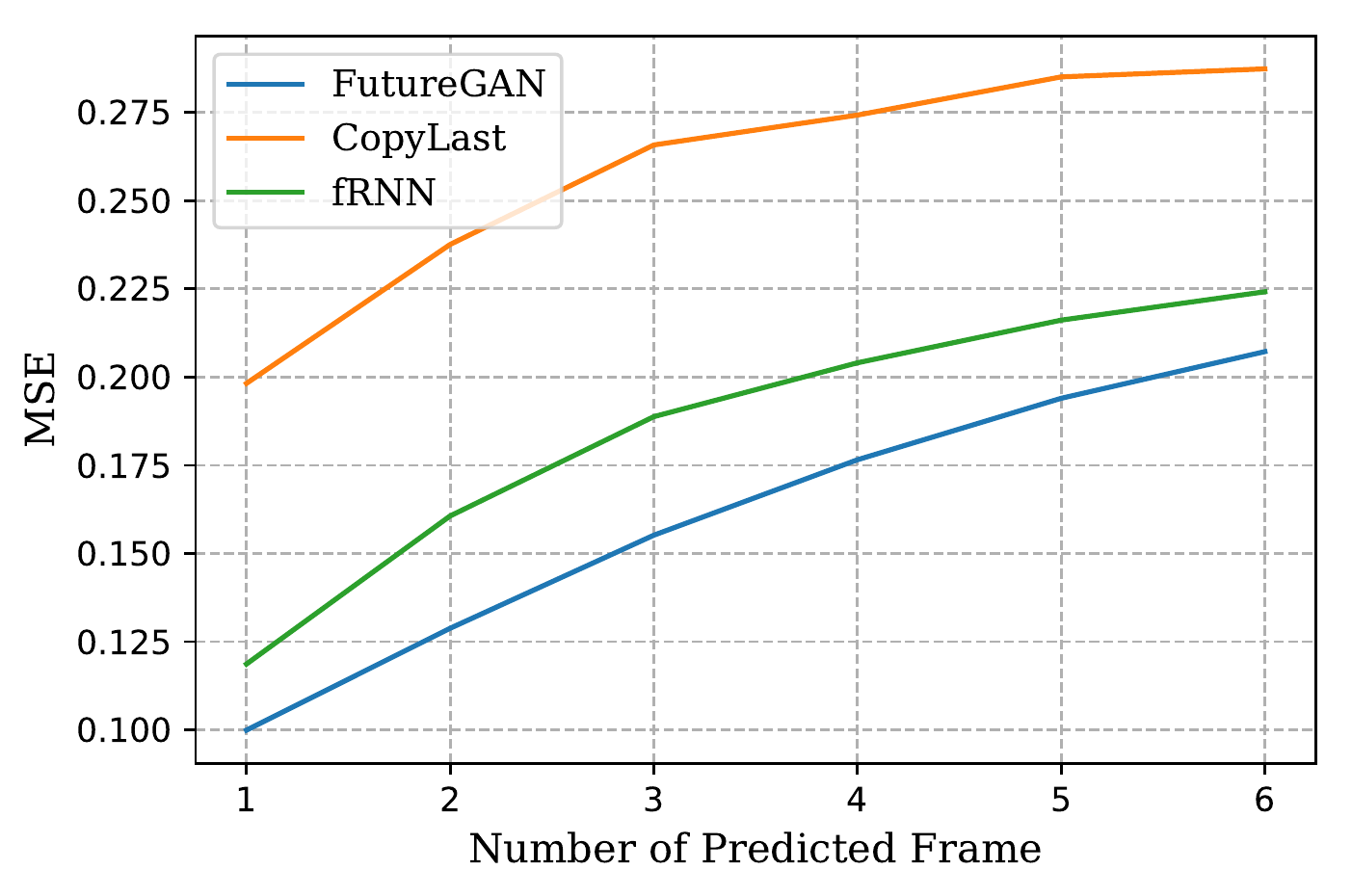}
		\label{movingmnist_mse}}
	\setcounter{subfigure}{0}
	\subfigure[MovingMNIST]{
		\includegraphics[width=0.3\linewidth]{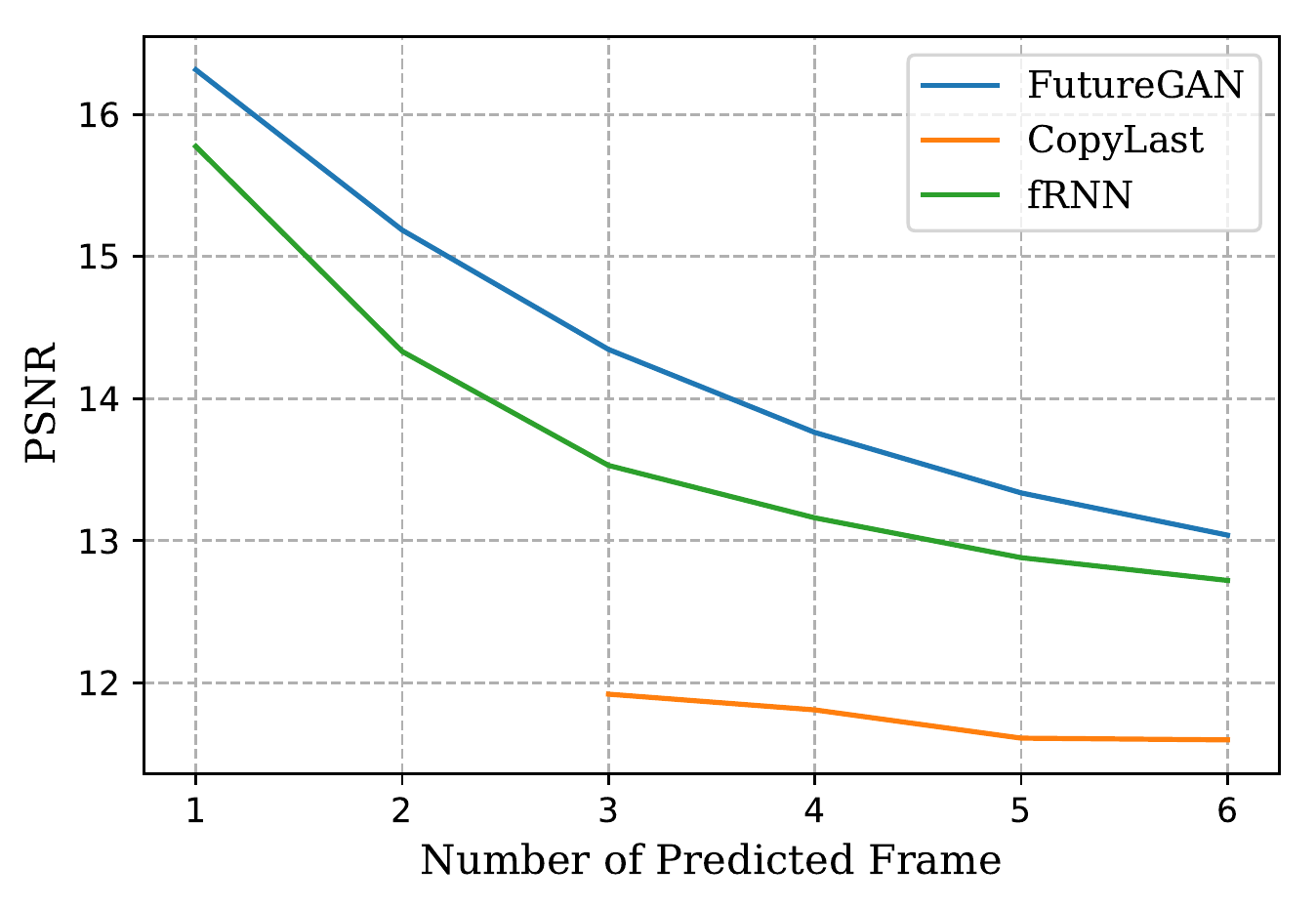}
		\label{movingmnist_psnr}}
	\subfigure{
		\includegraphics[width=0.3\linewidth]{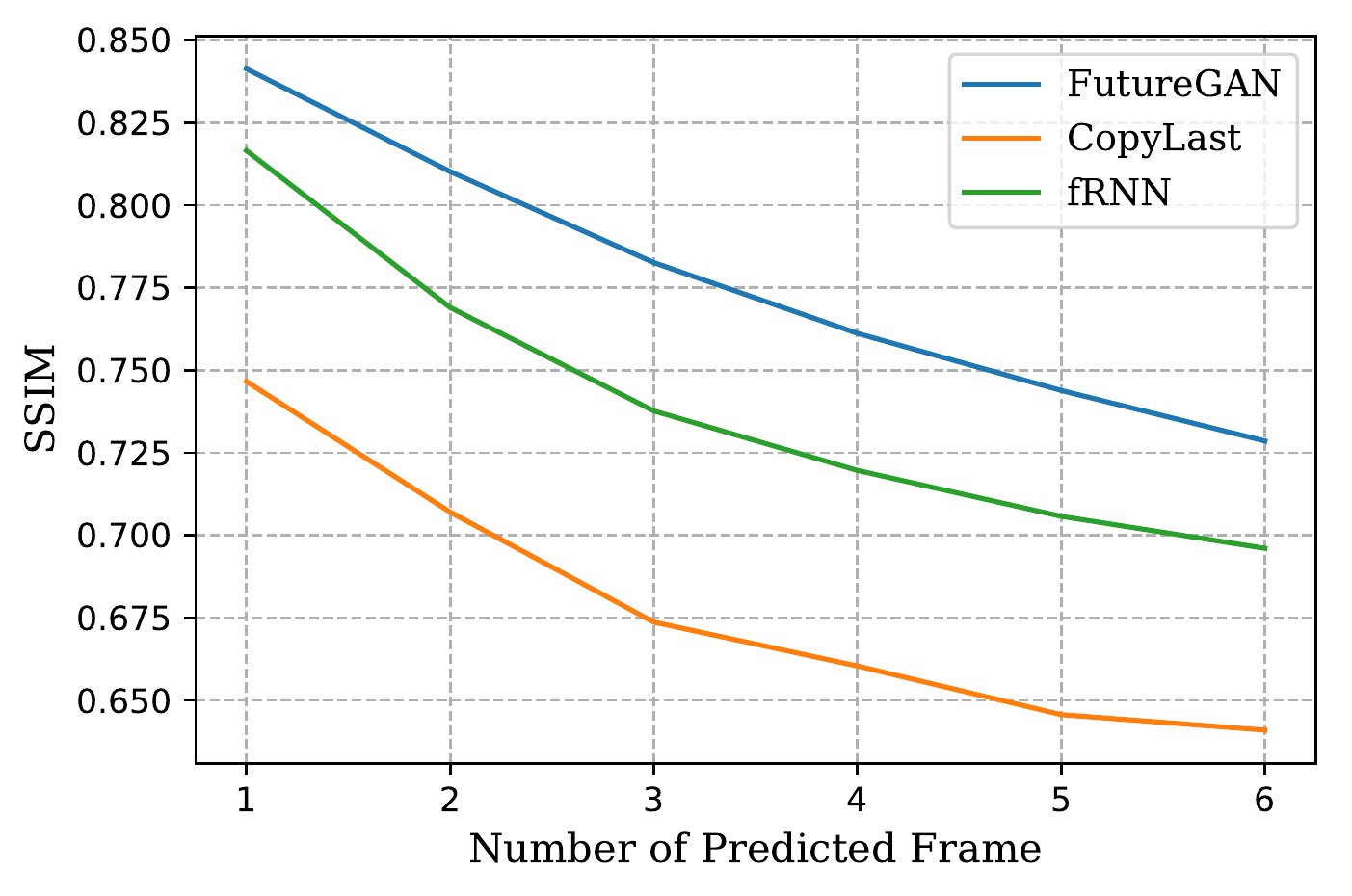}
		\label{movingmnist_ssim}}		
	\subfigure{
		\includegraphics[width=0.3\linewidth]{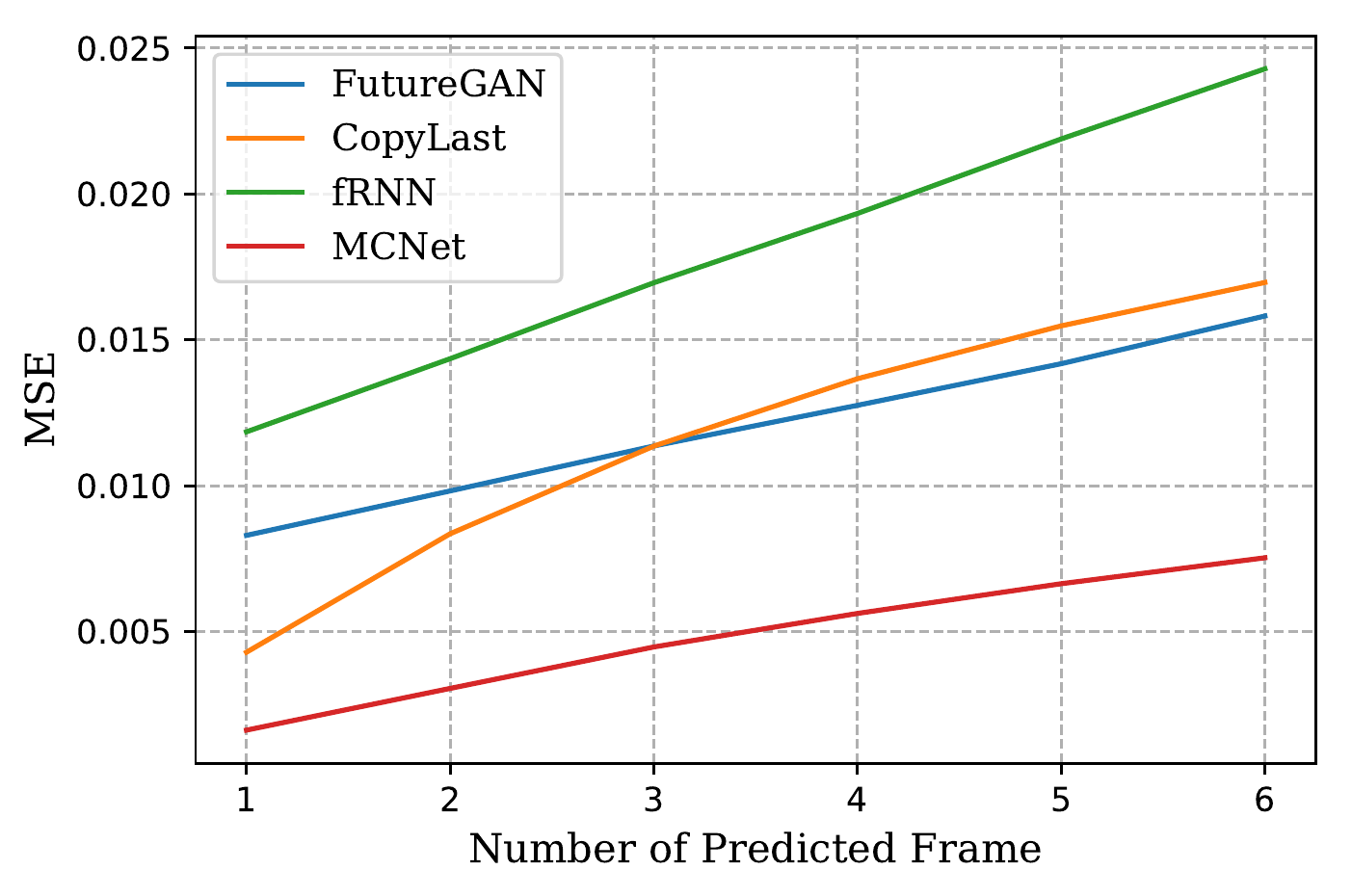}
		\label{kthaction_mse}}	
	\setcounter{subfigure}{1}	
	\subfigure[KTH Action]{
		\includegraphics[width=0.3\linewidth]{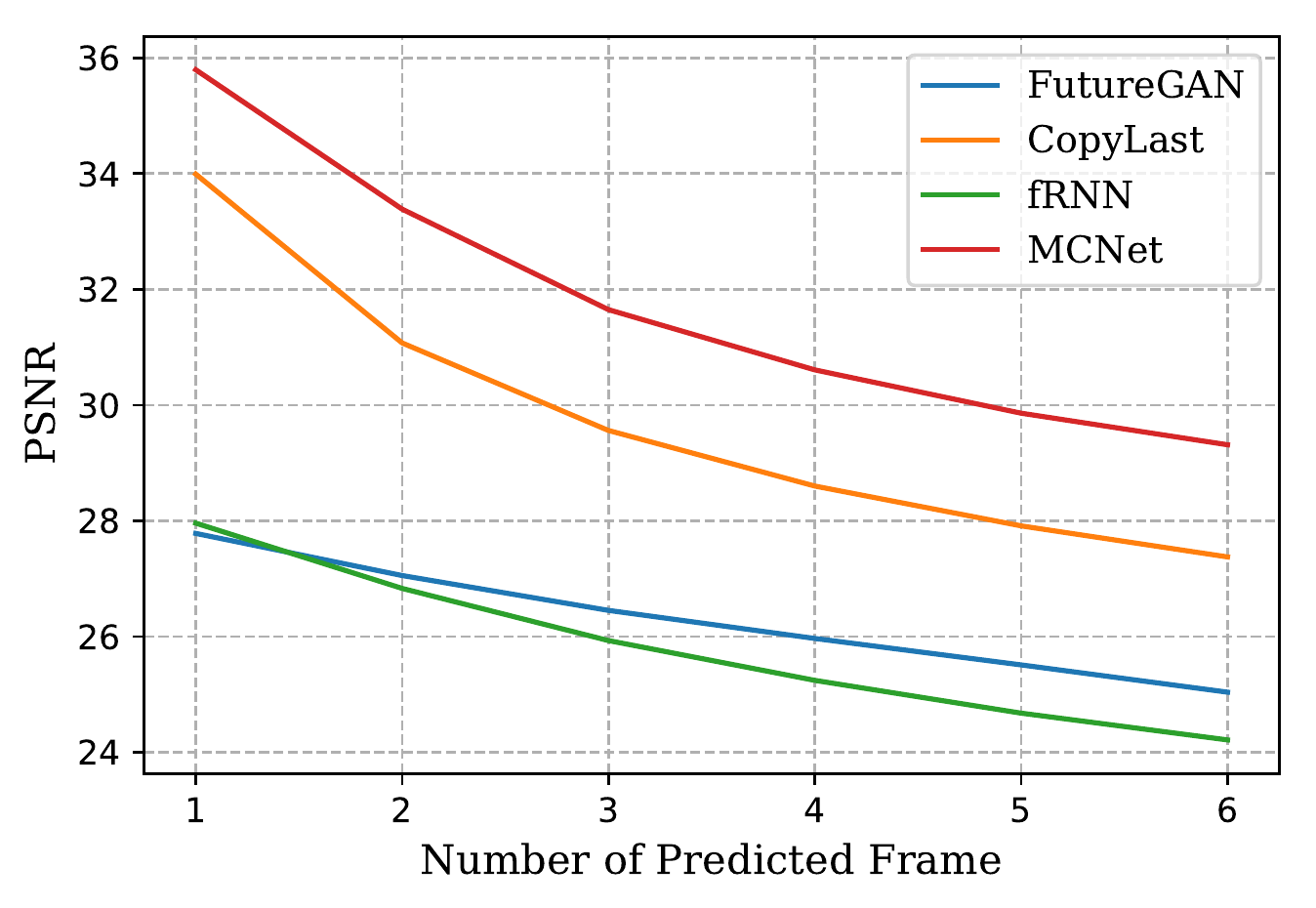}
		\label{kthaction_psnr}}
	\subfigure{
		\includegraphics[width=0.3\linewidth]{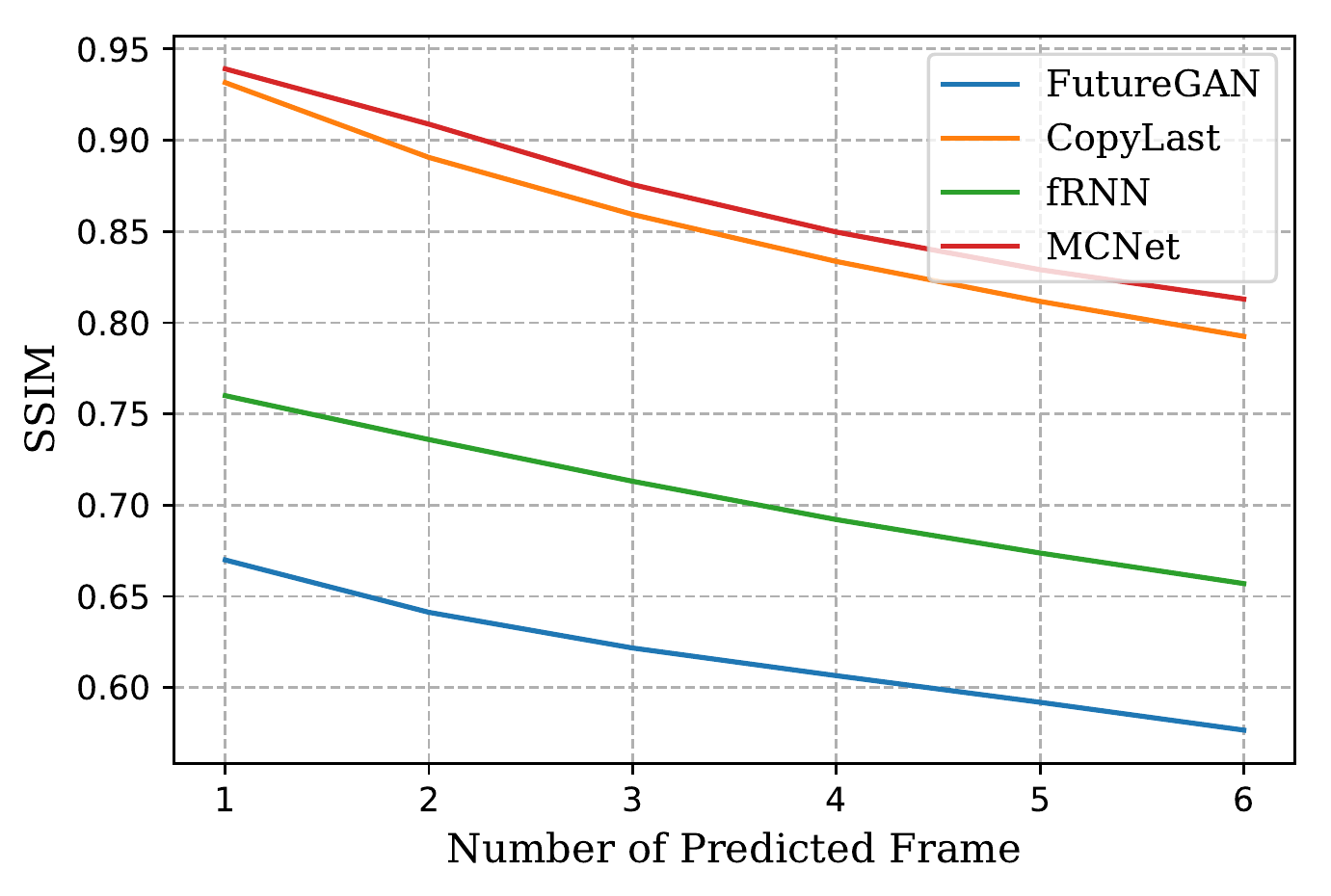}
		\label{kthaction_ssim}}	
	\caption{Quantitative results per predicted frame for the full test splits.}
	\label{quantitative}
\end{figure*}

\subsection{MovingMNIST}
To verify the effectiveness of our model architecture in general, we utilized the MovingMNIST dataset as a toy example. 
Following the procedure described in \citep{Srivastava2015}, we generated a set of 4500 videos for training, each of length 36 frames. 
Every MovingMNIST frame displays two white bouncing digits of distinct classes on a black background. 
Our generator network was trained to predict six future frames while being conditioned on six input frames, thus  a total of 13499 sequences was used for training. 
For testing, we generated another set of 2250 videos of length 36 frames, resulting in a test set containing 6750 sequences. 

\begin{figure}
	\centering
	\includegraphics[width=\linewidth]{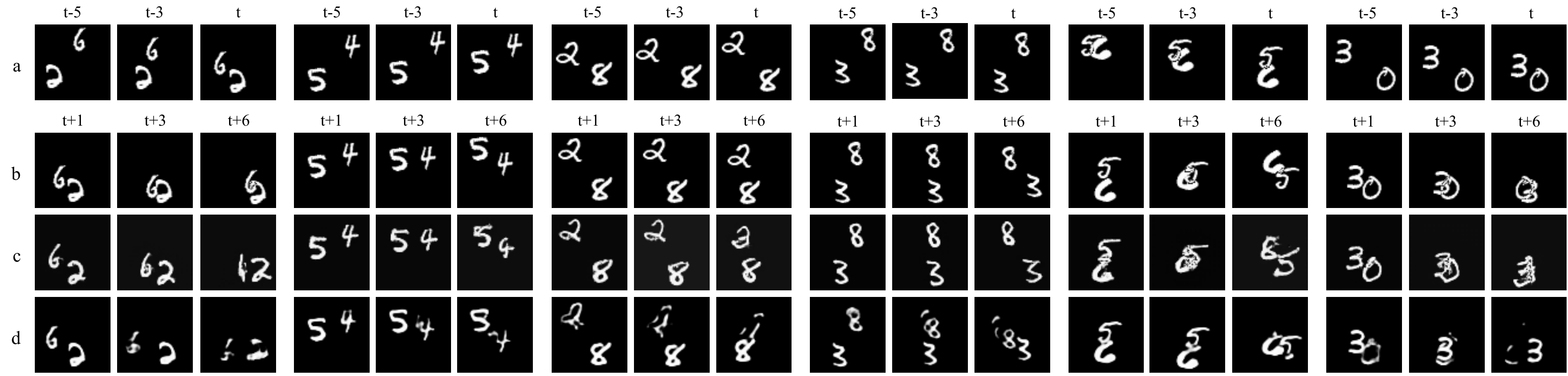}
	\caption{Prediction results for the MovingMNIST test sequences. a: Input, b: Ground Truth, c: FutureGAN (ours),  d: fRNN \citep{Oliu2018}}
	\label{movingmnist}
\end{figure}

In Figure \ref{movingmnist}, we show a qualitative comparison of our FutureGAN model to the \emph{fRNN} model of \citet{Oliu2018}.
The average quantitative results are listed in table \ref{movingmnist_kthaction_avg}. 
We provide the per frame values of the quantitative measures in figure \ref{quantitative}. 
Note that we used the pre-trained models provided by the original authors to generate the results.
This means, the fRNN was trained to predict 10 frames based on 10 input frames. 

\subsection{KTH Action}
In the second set of experiments, we used the KTH Action dataset. This dataset consists of 600 videos that display 25 different persons, each performing six actions in four different scenarios. 
The grayscale videos were recorded with a frame rate of \SI{25}{\fps} and have varying length. 
We split the dataset into person 1 to 16 for training, and 17 to 25 for testing. 
The FutureGAN model was trained to predict six future frames conditioned on six past frames. 
In total, our training set consists of 15156 sequences for predicting.  
Our test set had 8722 sequences. 

\begin{figure*}[h]
	\centering
	\includegraphics[width=\linewidth]{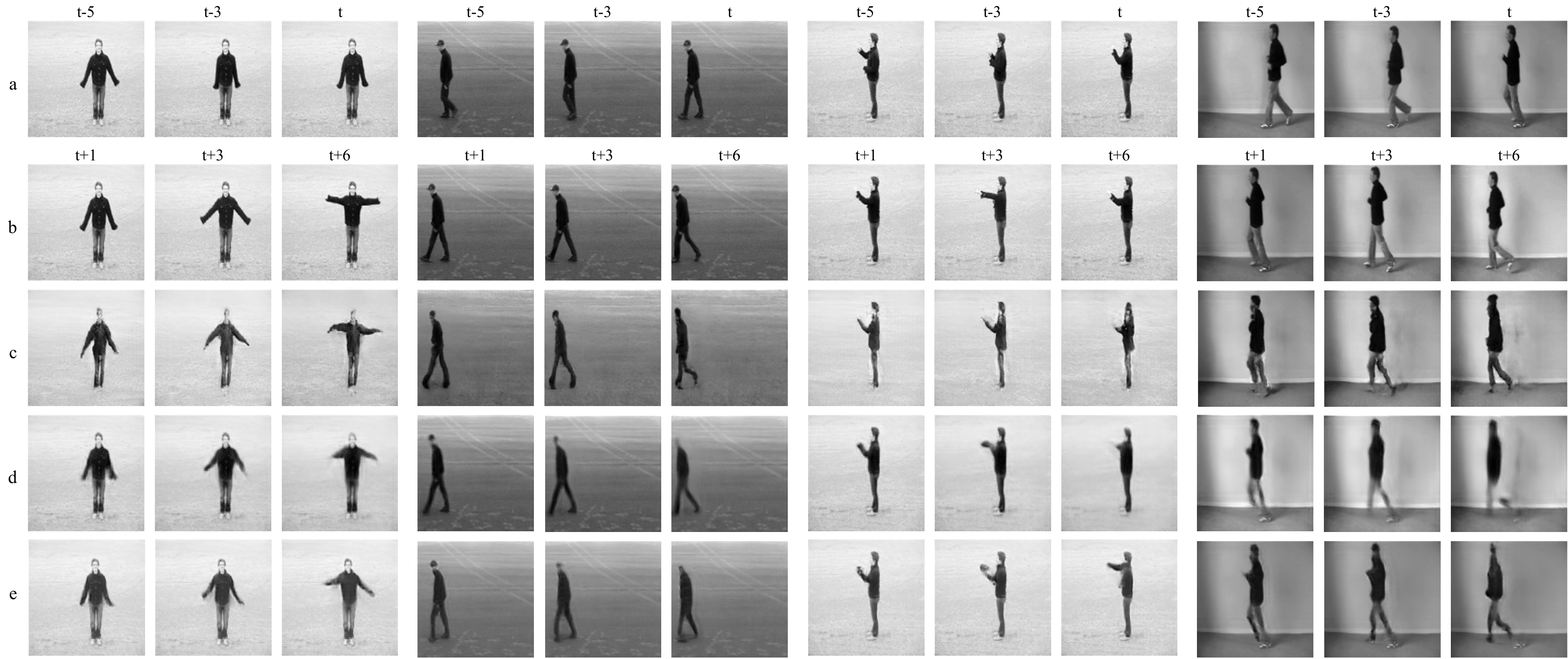}
	\caption{Prediction results for the KTH Action test split. a: Input, b: Ground Truth, c: FutureGAN (ours), d: fRNN \citep{Oliu2018}, e: MCNet \citep{Villegas2017}.}
	\label{kthaction}
\end{figure*}

In Figure \ref{kthaction}, we show a qualitative comparison of our FutureGAN model to the fRNN and the \emph{MCNet} of \citet{Villegas2017}. 
The average quantitative results are listed in table \ref{movingmnist_kthaction_avg}. 
We provide the per frame values of the quantitative measures in figure \ref{quantitative}.
For testing the MCNet, we used the pre-trained models provided by the original authors to generate the results.
The fRNN was originally trained for frame resolutions of \SI{80x64}{\px}. 
We re-trained the fRNN model on sequences with a \SI{128x128}{\px} frame resolution, following the procedure of \citep{Oliu2018}.
This means, both the fRNN and the MCNet model were trained to predict 10 frames based on 10 input frames. 

\subsection{Cityscapes}
To further investigate whether our model is able to scale to more complex real-world scenes, we trained it on the Cityscapes dataset. 
This dataset contains 2975 training videos and 1525 test videos, each of 30 frames in length. 
The \SI{16}{\bit} color videos were recorded with a frame rate of \SI{17}{\fps} in 50 different cities of Germany. 
We took the training and testing set as split by \citet{Cordts2016}. 
Each split contains the videos from a different set of cities.
The FutureGAN was trained to predict five future frames based on five input frames. 
In total, our training set consisted of 8924 frame sequences. 
The test set had 4574 sequences.

\begin{figure*}[h]
	\centering
	\includegraphics[width=0.9\linewidth]{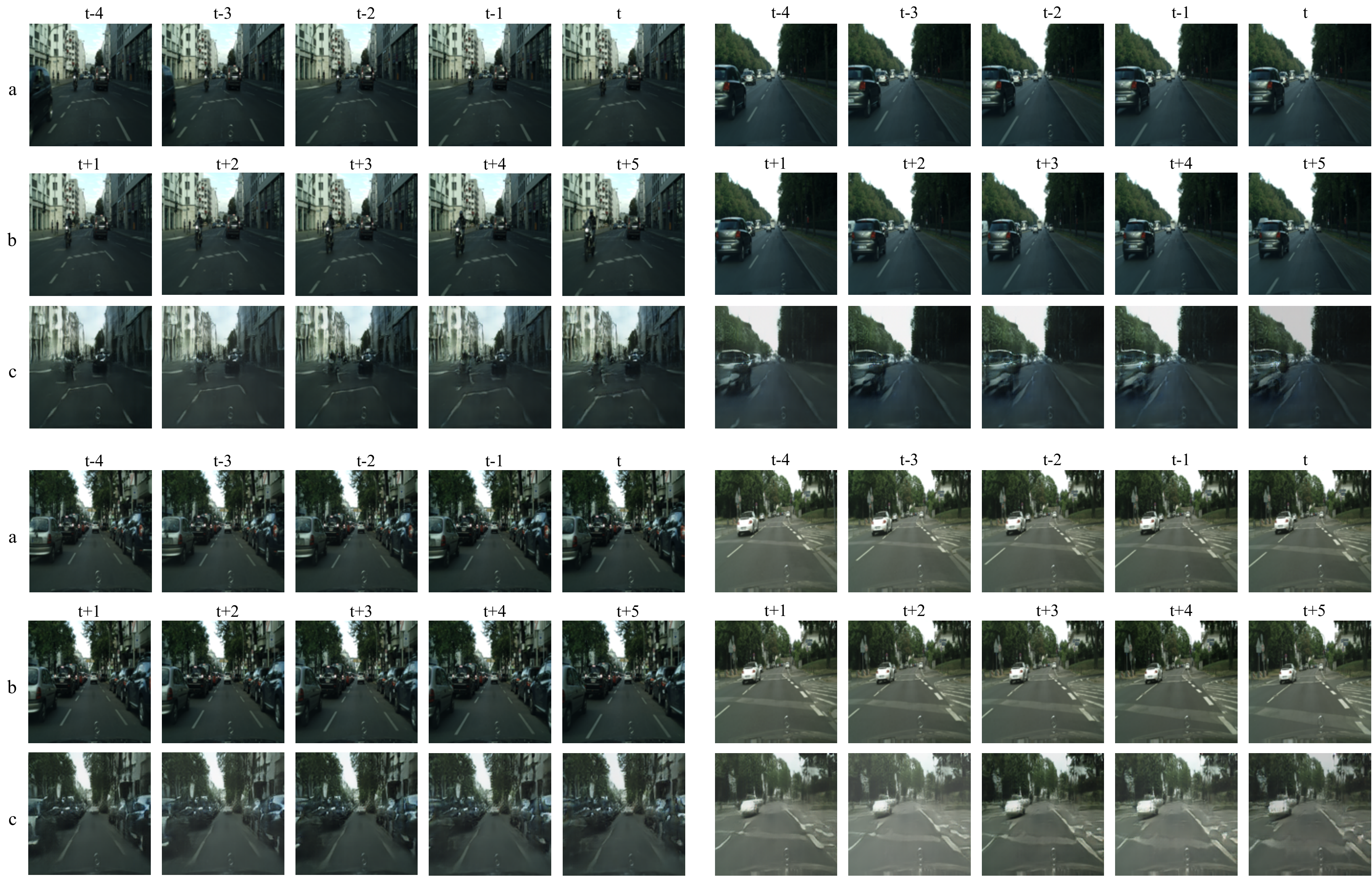}
	\caption{Prediction results for the Cityscapes test sequences. a: Input, b: Ground Truth, c: FutureGAN (ours)}
	\label{cityscapes}
\end{figure*}

In Figure \ref{cityscapes}, we display the qualitative results of our FutureGAN model trained on Cityscapes. 
For brevity, we provide the quantitative values in appendix \ref{appendix-cityscapes} figure \ref{cityscapes_quantitative}. 

\subsection{Long-term Predictions}
To test the generalization abilities of our network, we generated long-term predictions for all three datasets. 
This was achieved by feeding the predictions recursively back in as inputs. 

\begin{figure*}[h]
	\centering
	\includegraphics[width=\linewidth]{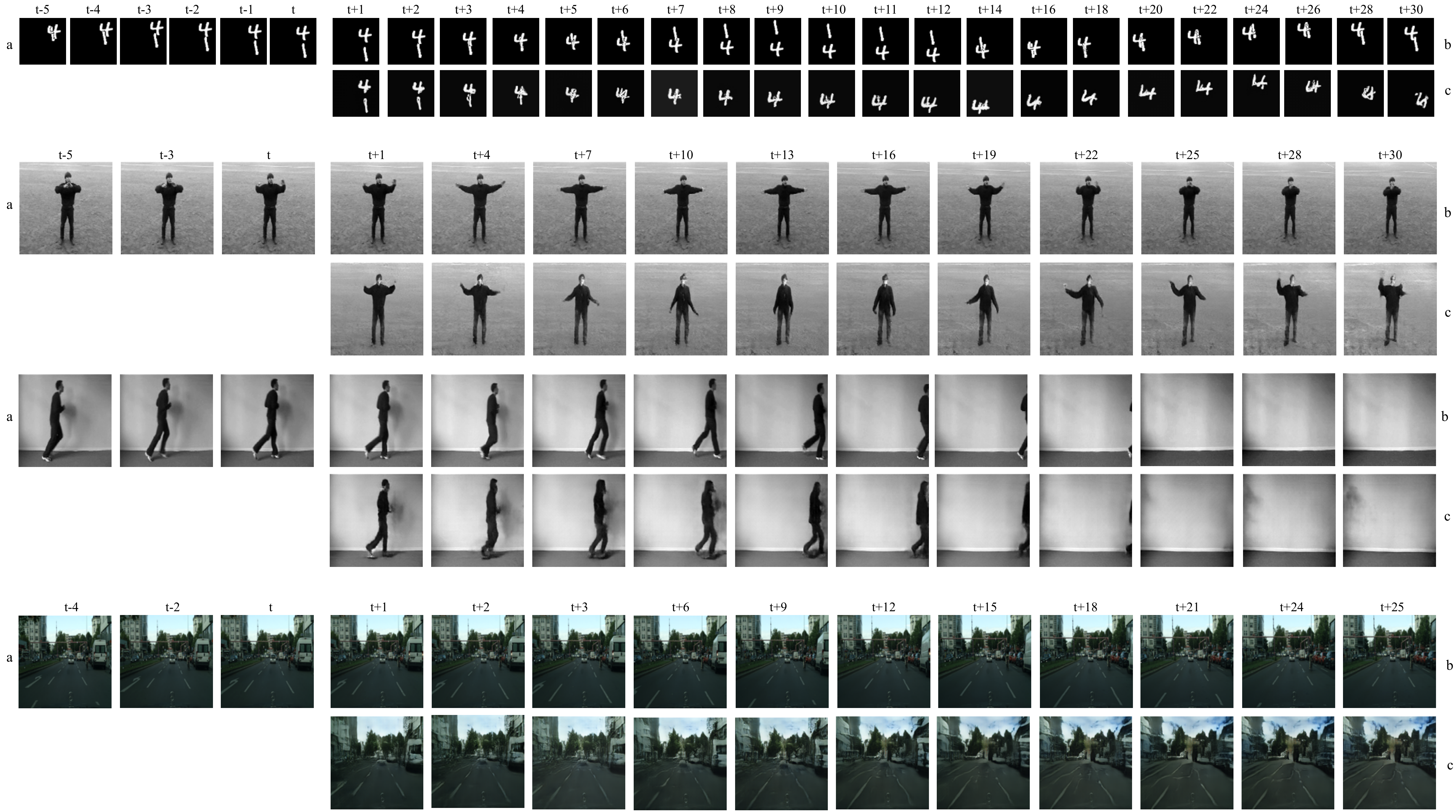}
	\caption{Prediction results of the long-term predictions for all three datasets. a: Input, b: Ground Truth, c: FutureGAN (ours)}
	\label{deeppred}
\end{figure*}

Figure \ref{deeppred} shows the qualitative results of this experiment. 
For MovingMNIST, we generated predictions for 30 frames ahead, letting the network observe only one real sequence of 6 input frames. 
On the KTH Action dataset, also predictions up to 30 frames ahead were made, while only one real sequence of 6 input frames was observed.
Additionally, we provide results for predictions up to 120 steps ahead for the KTH Action dataset in appendix \ref{appendix-kthaction_deeppred120}.
For Cityscapes, we generated predictions 25 frames ahead, when the network only observed one real sequence of 5 input frames. 

\section{Conclusion and Discussion}
In this paper, we have proposed FutureGAN, a new model that predicts future video frames conditioned on an input sequence. 
By extending the existing PGGAN architecture to video prediction, we are able to predict future frames that appear realistic, while the problems that typically arise when training GANs are avoided.
Our proposed model is trained to predict multiple future frames at once, using a similar setting for different datasets.
This makes FutureGAN directly applicable to a variety of datasets without utilizing dataset specific domain knowledge.
Contrary to other approaches, our networks solely use the raw pixel values as an input, without relying on additional priors, or conditional information. 

To evaluate our model, we trained and tested it on three datasets of increasing complexity. 
For MovingMNIST and KTH Action, we used an identical training setting, except for the dataset size and final frame resolution.
The predicted frames show that the network effectively learned representations of spatial and temporal transformations for the two datasets.
Our network identifies moving pixels in the input frames and transforms them based on its learned internal representations.
For both datasets, the results are competitive to the state-of-the-art.
The qualitative results of the Cityscapes dataset suggest that our model scales to complex natural traffic scenes as well. 
We observed that FutureGAN applies separate motion patterns to the background and foreground pixels. 
Furthermore, it seems as if the network was able to learn scene-specific representations of ego-motion. 
In most cases, it applies the correct motion patterns for either a static or dynamic background based on the input sequence. 
Even though the networks were trained on fewer frames, they generalize reasonably well to predict deeper into future for the KTH Action and Cityscapes dataset. 
The predictions still appear plausible, although the frames tend to get blurrier for increasing numbers of time steps.

Our experiments verify that the progressive growth strategy of \citet{Karras2018} scales effectively to the more complex video prediction task.
FutureGAN is a highly flexible model that can easily be trained on various datasets of different resolutions without prior knowledge about the data.

\subsubsection*{Acknowledgements}
The authors are very grateful for the computing resources provided by the Leibniz Supercomputing Centre (LRZ) of the Bavarian Academy of Sciences and Humanities (BAdW) and their excellent team, with special thanks to Yu Wang. 
Further, we gratefully acknowledge the support of NVIDIA Corporation with the donation of a Titan X Pascal GPU, used for this research. 
We thank Marc Oliu Simón for kindly providing the pre-trained fRNN models. 
Finally, we want to appreciate the valuable feedback of our colleague Lloyd Hughes and our colleagues of the Computer Vision Research Group.

\small
\setlength{\bibsep}{0.0pt}
\bibliographystyle{plainnat}
\bibliography{./Bib/Bib}

\begin{thebibliography}{48}
\providecommand{\natexlab}[1]{#1}
\providecommand{\url}[1]{\texttt{#1}}
\expandafter\ifx\csname urlstyle\endcsname\relax
  \providecommand{\doi}[1]{doi: #1}\else
  \providecommand{\doi}{doi: \begingroup \urlstyle{rm}\Url}\fi

\bibitem[Arjovsky et~al.(2018)Arjovsky, Chitala, and Bottou]{Arjovsky2017}
M.~Arjovsky, S.~Chitala, and L.~Bottou.
\newblock {W}asserstein {GAN}.
\newblock In \emph{{ICML}}, 2018.

\bibitem[Babaeizadeh et~al.(2018)Babaeizadeh, Finn, Erhan, Campbell, and
  Levine]{Babaeizadeh2018}
M.~Babaeizadeh, C.~Finn, D.~Erhan, R.~H. Campbell, and S.~Levine.
\newblock {S}tochastic {V}ariational {V}ideo {P}rediction.
\newblock In \emph{{ICLR}}, 2018.

\bibitem[Bhattacharjee and Das(2017)]{Bhattacharjee2017}
P.~Bhattacharjee and S.~Das.
\newblock {T}emporal {C}oherency based {C}riteria for {P}redicting {V}ideo
  {F}rames using {D}eep {M}ulti-stage {G}enerative {A}dversarial {N}etworks.
\newblock In \emph{{NIPS}}. Curran Associates, Inc., 2017.

\bibitem[Byeon et~al.(2018)Byeon, Wang, Srivastava, and
  Koumoutsakos]{Byeon2018}
W.~Byeon, Q.~Wang, R.~K. Srivastava, and P.~Koumoutsakos.
\newblock {F}ully {C}ontext-{A}ware {V}ideo {P}rediction.
\newblock In \emph{{ECCV}}, 2018.

\bibitem[Chen et~al.(2017)Chen, Wang, Wang, and Chen]{Chen2017}
B.~Chen, W.~Wang, J.~Wang, and X.~Chen.
\newblock {V}ideo {I}magination from a {S}ingle {I}mage with {T}ransformation
  {G}eneration.
\newblock In \emph{{MM}}, 2017.

\bibitem[Cordts et~al.(2016)Cordts, Omran, Ramos, Rehfeld, Enzweiler, Benenson,
  Franke, Roth, and Schiele]{Cordts2016}
M.~Cordts, M.~Omran, S.~Ramos, T.~Rehfeld, M.~Enzweiler, R.~Benenson,
  U.~Franke, S.~Roth, and B.~Schiele.
\newblock {T}he {C}ityscapes {D}ataset for {S}emantic {U}rban {S}cene
  {U}nderstanding.
\newblock In \emph{{CVPR}}, 2016.

\bibitem[De~Brabandere et~al.(2016)De~Brabandere, Jia, Tuytelaars, and
  Van~Gool]{DeBrabandere2016}
B.~De~Brabandere, X.~Jia, T.~Tuytelaars, and L.~Van~Gool.
\newblock {D}ynamic {F}ilter {N}etworks.
\newblock In \emph{{NIPS}}. Curran Associates, Inc., 2016.

\bibitem[Denton and Fergus(2018)]{Denton2018}
E.~Denton and R.~Fergus.
\newblock {S}tochastic {V}ideo {G}eneration with a {L}earned {P}rior.
\newblock In \emph{{ICML}}, 2018.

\bibitem[Denton and Birodkar(2017)]{Denton2017}
E.~L. Denton and V.~Birodkar.
\newblock {U}nsupervised {L}earning of {D}isentangled {R}epresentations from
  {V}ideo.
\newblock In \emph{{NIPS}}. Curran Associates, Inc., 2017.

\bibitem[Farnebäck(2003)]{Farnebaeck2003}
G.~Farnebäck.
\newblock {T}wo-{F}rame {M}otion {E}stimation {B}ased on {P}olynomial
  {E}xpansion.
\newblock In \emph{{SCIA}}, 2003.

\bibitem[Finn et~al.(2016)Finn, Goodfellow, and Levine]{Finn2016}
C.~Finn, I.~Goodfellow, and S.~Levine.
\newblock {U}nsupervised {L}earning for {P}hysical {I}nteraction through
  {V}ideo {P}rediction.
\newblock In \emph{{NIPS}}. Curran Associates, Inc., 2016.

\bibitem[Goodfellow et~al.(2014)Goodfellow, Pouget-Abadie, Mirza, Xu,
  Warde-Farley, Ozair, Courville, and Bengio]{Goodfellow2014}
I.~Goodfellow, J.~Pouget-Abadie, M.~Mirza, B.~Xu, D.~Warde-Farley, S.~Ozair,
  A.~Courville, and Y.~Bengio.
\newblock {G}enerative {A}dversarial {N}etworks.
\newblock In \emph{{NIPS}}. Curran Associates, Inc., 2014.

\bibitem[Goroshin et~al.(2015)Goroshin, Mathieu, and LeCun]{Goroshin2015}
R.~Goroshin, M.~Mathieu, and Y.~LeCun.
\newblock {L}earning to {L}inearize {U}nder {U}ncertainty.
\newblock In \emph{{NIPS}}. Curran Associates, Inc., 2015.

\bibitem[Gulrajani et~al.(2017)Gulrajani, Ahmed, Arjovsky, Dumoulin, and
  Courville]{Gulrajani2017}
I.~Gulrajani, F.~Ahmed, M.~Arjovsky, V.~Dumoulin, and A.~Courville.
\newblock {I}mproved {T}raining of {W}asserstein {GAN}s.
\newblock In \emph{{NIPS}}. Curran Associates, Inc., 2017.

\bibitem[Hao et~al.(2018)Hao, Huang, and Belongie]{Hao2018}
Z.~Hao, X.~Huang, and S.~Belongie.
\newblock {C}ontrollable {V}ideo {G}eneration with {S}parse {T}rajectories.
\newblock In \emph{{CVPR}}, 2018.

\bibitem[He et~al.(2015)He, Zhang, Ren, and Sun]{He2015}
K.~He, X.~Zhang, S.~Ren, and J.~Sun.
\newblock {D}elving {D}eep into {R}ectifiers: {S}urpassing {H}uman-{L}evel
  {P}erformance on {I}mage{N}et {C}lassification.
\newblock In \emph{{ICCV}}, 2015.

\bibitem[Kalchbrenner et~al.(2017)Kalchbrenner, van~den Oord, Simonyan,
  Danihelka, Vinyals, Graves, and Kavukcuoglu]{Kalchbrenner2017}
N.~Kalchbrenner, A.~van~den Oord, K.~Simonyan, I.~Danihelka, O.~Vinyals,
  A.~Graves, and K.~Kavukcuoglu.
\newblock {V}ideo {P}ixel {N}etworks.
\newblock In \emph{{ICML}}, 2017.

\bibitem[Karras et~al.(2018)Karras, Aila, Laine, and Lehtinen]{Karras2018}
T.~Karras, T.~Aila, S.~Laine, and J.~Lehtinen.
\newblock {P}rogressive {G}rowing of {GAN}s for {I}mproved {Q}uality,
  {S}tability, and {V}ariation.
\newblock In \emph{{ICLR}}, 2018.

\bibitem[Kingma and Ba(2015)]{Kingma2015}
D.~P. Kingma and J.~Ba.
\newblock {A}dam: {A} {M}ethod for {S}tochastic {O}ptimization.
\newblock In \emph{{ICLR}}, 2015.

\bibitem[Kratzwald et~al.(2017)Kratzwald, Huang, Paudel, , Dinesh, and
  Van~Gool]{Kratzwald2017}
B.~Kratzwald, Z.~Huang, D.~P. Paudel, , A.~Dinesh, and L.~Van~Gool.
\newblock {I}mproving {V}ideo {G}eneration for {M}ulti-functional
  {A}pplications.
\newblock \emph{{C}o{RR}}, abs/1711.11453, 2017.

\bibitem[Krizhevsky et~al.(2012)Krizhevsky, Sutskever, and
  Hinton]{Krizhevsky2012}
A.~Krizhevsky, I.~Sutskever, and G.~E. Hinton.
\newblock {I}mage{N}et {C}lassification with {D}eep {C}onvolutional {N}eural
  {N}etworks.
\newblock In \emph{{NIPS}}. Curran Associates, Inc., 2012.

\bibitem[Lee et~al.(2018)Lee, Zhang, Ebert, Abbeel, Finn, and Levine]{Lee2018}
A.~X. Lee, R.~Zhang, F.~Ebert, P.~Abbeel, C.~Finn, and S.~Levine.
\newblock {S}tochastic {A}dversarial {V}ideo {P}rediction.
\newblock \emph{{C}o{RR}}, abs/1804.01523, 2018.

\bibitem[Liang et~al.(2017)Liang, Lee, Dai, and Xing]{Liang2017}
X.~Liang, L.~Lee, W.~Dai, and E.~P. Xing.
\newblock {D}ual {M}otion {GAN} for {F}uture-{F}low {E}mbedded {V}ideo
  {P}rediction.
\newblock In \emph{{ICCV}}, 2017.

\bibitem[Liu et~al.(2017)Liu, Yeh, Tang, Liu, and Agarwala]{Liu2017}
Z.~Liu, R.~A. Yeh, X.~Tang, Y.~Liu, and A.~Agarwala.
\newblock {V}ideo {F}rame {S}ynthesis {U}sing {D}eep {V}oxel {F}low.
\newblock In \emph{{ICCV}}, 2017.

\bibitem[Lotter et~al.(2016)Lotter, Kreiman, and Cox]{Lotter2016}
W.~Lotter, G.~Kreiman, and D.~Cox.
\newblock {U}nsupervised {L}earning of {V}isual {S}tructure using {P}redictive
  {G}enerative {N}etworks.
\newblock In \emph{{ICLR}}, 2016.

\bibitem[Lotter et~al.(2017)Lotter, Kreiman, and Cox]{Lotter2017}
W.~Lotter, G.~Kreiman, and D.~Cox.
\newblock {D}eep {P}redictive {C}oding {N}etworks for {V}ideo {P}rediction and
  {U}nsupervised {L}earning.
\newblock In \emph{{ICLR}}, 2017.

\bibitem[Lu et~al.(2017)Lu, Hirsch, and Schlkopf]{Lu2017}
C.~Lu, M.~Hirsch, and B.~Schlkopf.
\newblock {F}lexible {S}patio-{T}emporal {N}etworks for {V}ideo {P}rediction.
\newblock In \emph{{CVPR}}, 2017.

\bibitem[Mahjourian et~al.(2017)Mahjourian, Wicke, and
  Angelova]{Mahjourian2017}
R.~Mahjourian, M.~Wicke, and A.~Angelova.
\newblock {G}eometry-{B}ased {N}ext {F}rame {P}rediction from {M}onocular
  {V}ideo.
\newblock In \emph{{IV}}, 2017.

\bibitem[Mathieu et~al.(2016)Mathieu, Couprie, and LeCun]{Mathieu2016}
M.~Mathieu, C.~Couprie, and Y.~LeCun.
\newblock {D}eep multi-scale video prediction beyond mean square error.
\newblock In \emph{{ICLR}}, 2016.

\bibitem[Oh et~al.(2015)Oh, Guo, Lee, Lewis, and Singh]{Oh2015}
J.~Oh, X.~Guo, H.~Lee, R.~L. Lewis, and S.~Singh.
\newblock {A}ction-{C}onditional {V}ideo {P}rediction using {D}eep {N}etworks
  in {A}tari {G}ames.
\newblock In \emph{{NIPS}}. Curran Associates, Inc., 2015.

\bibitem[Oliu et~al.(2018)Oliu, Selva, and Escalera]{Oliu2018}
M.~Oliu, J.~Selva, and S.~Escalera.
\newblock {F}olded {R}ecurrent {N}eural {N}etworks for {F}uture {V}ideo
  {P}rediction.
\newblock In \emph{{ECCV}}, 2018.

\bibitem[Patraucean et~al.(2016)Patraucean, Handa, and Cipolla]{Patraucean2016}
V.~Patraucean, A.~Handa, and R.~Cipolla.
\newblock {S}patio-{T}emporal {V}ideo {A}utoencoder with {D}ifferentiable
  {M}emory.
\newblock In \emph{{ICLR}}, 2016.

\bibitem[Ranzato et~al.(2014)Ranzato, Szlam, Bruna, Mathieu, Collobert, and
  Chopra]{Ranzato2014}
M.'A. Ranzato, A.~Szlam, J.~Bruna, M.~Mathieu, R.~Collobert, and S.~Chopra.
\newblock {V}ideo ({L}anguage) {M}odeling: {A} {B}aseline for generative
  {M}odels of natural {V}ideos.
\newblock \emph{{C}o{RR}}, abs/1412.6604, 2014.

\bibitem[Saito et~al.(2017)Saito, Matsumoto, and Saito]{Saito2017}
M.~Saito, E.~Matsumoto, and S.~Saito.
\newblock {T}emporal {G}enerative {A}dversarial {N}ets {W}ith {S}ingular
  {V}alue {C}lipping.
\newblock In \emph{{ICCV}}, 2017.

\bibitem[Salimans et~al.(2016)Salimans, Goodfellow, Zaremba, Cheung, Radford,
  and Chen]{Salimans2016}
T.~Salimans, I.~J. Goodfellow, W.~Zaremba, V.~Cheung, A.~Radford, and X.~Chen.
\newblock {I}mproved {T}echniques for {T}raining {GAN}s.
\newblock In \emph{{NIPS}}. Curran Associates, Inc., 2016.

\bibitem[Schüldt et~al.(2004)Schüldt, Laptev, and Caputo]{Schuldt2004}
C.~Schüldt, I.~Laptev, and B.~Caputo.
\newblock {R}ecognizing {H}uman {AA}ction: {A} {L}ocal {SVM} {A}pproach.
\newblock In \emph{{ICPR}}, 2004.

\bibitem[Srivastava et~al.(2015)Srivastava, Mansimov, and
  Salakhudinov]{Srivastava2015}
N.~Srivastava, E.~Mansimov, and R.~Salakhudinov.
\newblock {U}nsupervised {L}earning of {V}ideo {R}epresentation using {LSTM}s.
\newblock In \emph{{ICML}}, pages 843--852, 2015.

\bibitem[Tran et~al.(2015)Tran, Bourdev, Fergus, Torresani, and
  Paluri]{Tran2015}
D.~Tran, L.~Bourdev, R.~Fergus, L.~Torresani, and M.~Paluri.
\newblock {L}earning {S}patiotemporal {F}eatures with 3{D} {C}onvolutional
  {N}etworks.
\newblock In \emph{{ICCV}}, 2015.

\bibitem[Tulyakov et~al.(2018)Tulyakov, Liu, Yang, and Kautz]{Tulyakov2018}
S.~Tulyakov, M.-Y. Liu, X.~Yang, and J.~Kautz.
\newblock {M}o{C}o{GAN}: {D}ecomposing {M}otion and {C}ontent for {V}ideo
  {G}eneration.
\newblock In \emph{{CVPR}}, 2018.

\bibitem[Villegas et~al.(2017)Villegas, Yang, Hong, Lin, and Lee]{Villegas2017}
R.~Villegas, J.~Yang, S.~Hong, X.~Lin, and H.~Lee.
\newblock {D}ecomposing {M}otion and {C}ontent for {N}atural {V}ideo {S}equence
  {P}rediction.
\newblock In \emph{{ICLR}}, 2017.

\bibitem[Vondrick et~al.(2016)Vondrick, Pirsiavash, and
  Torralba]{Vondrick2016b}
C.~Vondrick, H.~Pirsiavash, and A.~Torralba.
\newblock {G}enerating {V}ideos with {S}cene {D}ynamics.
\newblock In \emph{{NIPS}}. Curran Associates, Inc., 2016.

\bibitem[Vondrick et~al.(2017)Vondrick, Pirsiavash, and Torralba]{Vondrick2017}
C.~Vondrick, H.~Pirsiavash, and A.~Torralba.
\newblock {G}enerating the {F}uture with {A}dversarial {T}ransformers.
\newblock In \emph{{CVPR}}, 2017.

\bibitem[Vukoti et~al.(2016)Vukoti, Pintea, Raymond, Gravier, and
  Van~Gemert]{Vukotic2016}
V.~Vukoti, A.-L. Pintea, C.~Raymond, G.~Gravier, and J.~Van~Gemert.
\newblock {O}ne-{S}tep {T}ime-{D}ependent {F}uture {V}ideo {F}rame {P}rediction
  with a {C}onvolutional {E}ncoder-{D}ecoder {N}eural {N}etwork.
\newblock In \emph{{NCCV}}, 2016.

\bibitem[Walker et~al.(2016)Walker, Doersch, Gupta, and Hebert]{Walker2016}
J.~Walker, C.~Doersch, A.~Gupta, and M.~Hebert.
\newblock {A}n {U}ncertain {F}uture: {F}orecasting from {S}tatic {I}mages using
  {V}ariational {A}utoencoders.
\newblock In \emph{{ECCV}}, 2016.

\bibitem[Wang et~al.(2017)Wang, Long, Wang, Gao, and Yu]{Wang2017}
Y.~Wang, M.~Long, J.~Wang, Z.~Gao, and P.~S. Yu.
\newblock {P}red{RNN}: {R}ecurrent {N}eural {N}etworks for {P}redictive
  {L}earning using {S}patiotemporal {LSTM}s.
\newblock In \emph{{NIPS}}. Curran Associates, Inc., 2017.

\bibitem[Xiong et~al.(2018)Xiong, Luo, Ma, Liu, and Luo]{Xiong2018}
W.~Xiong, W.~Luo, L.~Ma, W.~Liu, and J.~Luo.
\newblock {L}earning to {G}enerate {T}ime-{L}apse {V}ideos {U}sing
  {M}ulti-{S}tage {D}ynamic {G}enerative {A}dversarial {N}etworks.
\newblock In \emph{{CVPR}}, 2018.

\bibitem[Xue et~al.(2016)Xue, Wu, Bouman, and Freeman]{Xue2016}
T.~Xue, J.~Wu, K.~L. Bouman, and W.~T. Freeman.
\newblock {V}isual {D}ynamics: {P}robabilistic {F}uture {F}rame {S}ynthesis via
  {C}ross {C}onvolutional {N}etworks.
\newblock In \emph{{NIPS}}. Curran Associates, Inc., 2016.

\bibitem[Zeng et~al.(2017)Zeng, Shen, Huang, Sun, and Niebles]{Zeng2017}
K.-H. Zeng, W.~B. Shen, D.-A. Huang, M.~Sun, and J.~C. Niebles.
\newblock {V}isual {F}orecasting by {I}mitating {D}ynamics in {N}atural
  {S}equences.
\newblock In \emph{{ICCV}}, 2017.

\end{thebibliography}

\newpage
\part*{Appendix}
\setcounter{section}{0}
\renewcommand{\thesection}{\Alph{section}}
\setcounter{table}{0}
\renewcommand{\thetable}{A.\arabic{table}}
\setcounter{figure}{0}
\renewcommand{\thefigure}{A.\arabic{figure}}

\section{Optical Flow Evaluation} \label{appendix-opticalflow}
We computed the dense optical flow maps using the Gunnar Farneback's algorithm \cite{Farnebaeck2003}.

\subsection{MovingMNIST Predictions}
\begin{figure}[h]
	\centering
	\includegraphics[width=\linewidth]{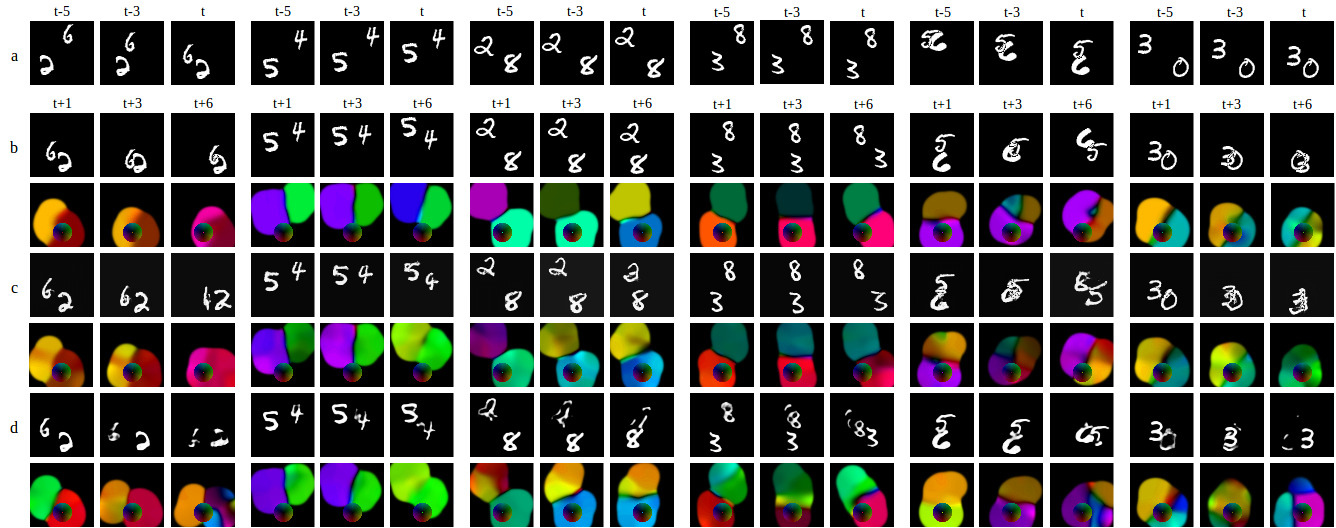}
	\caption{Prediction results for the MovingMNIST test sequences. a: Input, b: Ground Truth, c: FutureGAN (ours),  d: fRNN \citep{Oliu2018}}
	\label{movingmnist_opticalflow}
\end{figure}

\subsection{KTHAction Predictions}
\begin{figure}[h]
	\centering
	\includegraphics[width=0.95\linewidth]{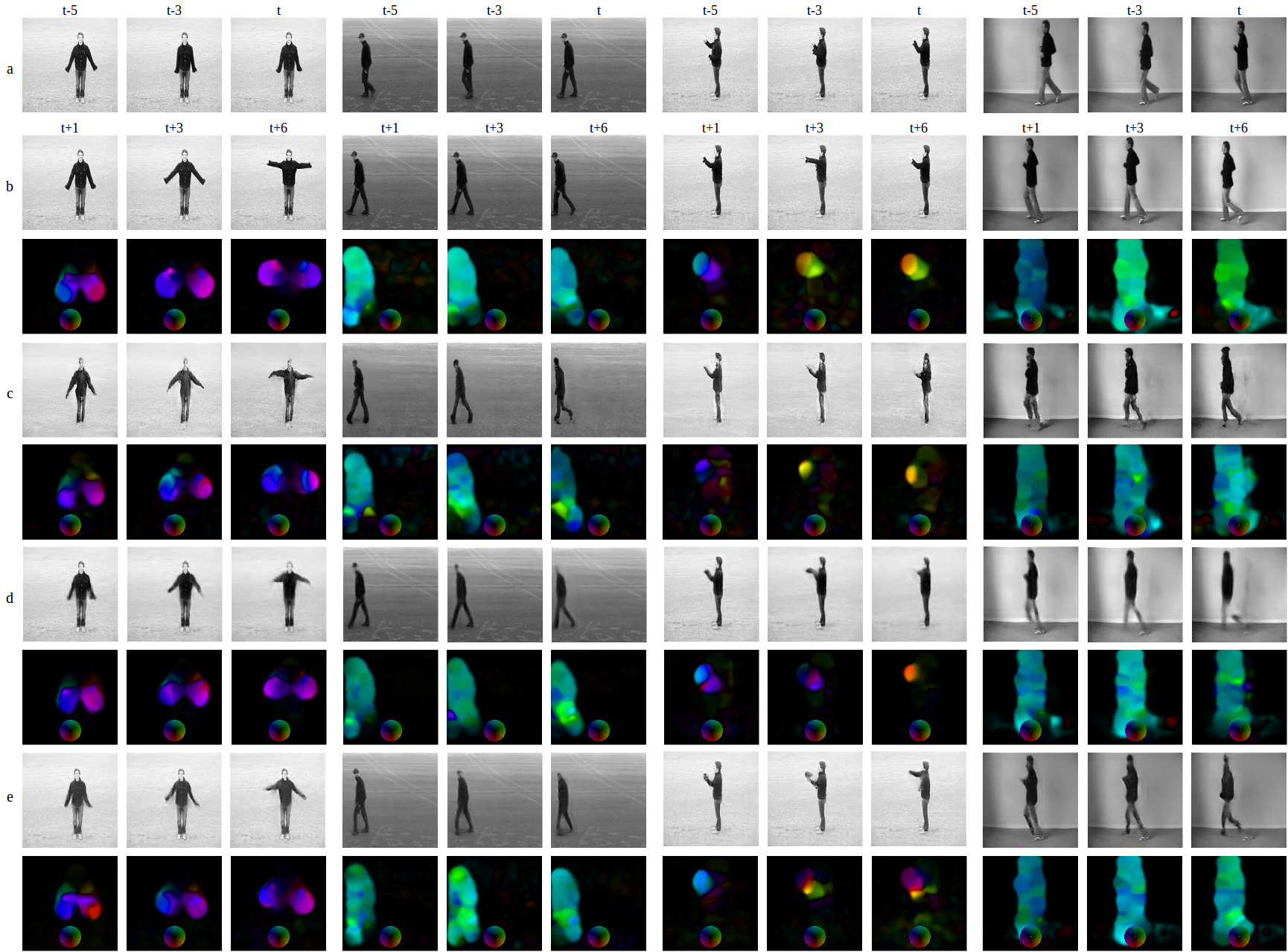}
	\caption{Prediction results for the KTH Action test split. a: Input, b: Ground Truth, c: FutureGAN (ours), d: fRNN \citep{Oliu2018}, e: MCNet \citep{Villegas2017}.}
	\label{kthaction_opticalflow}
\end{figure}

\newpage
\subsection{KTHAction 120 step Long-Term Predictions} \label{appendix-kthaction_deeppred120}
\begin{figure}[h]
	\centering
	\includegraphics[width=\linewidth]{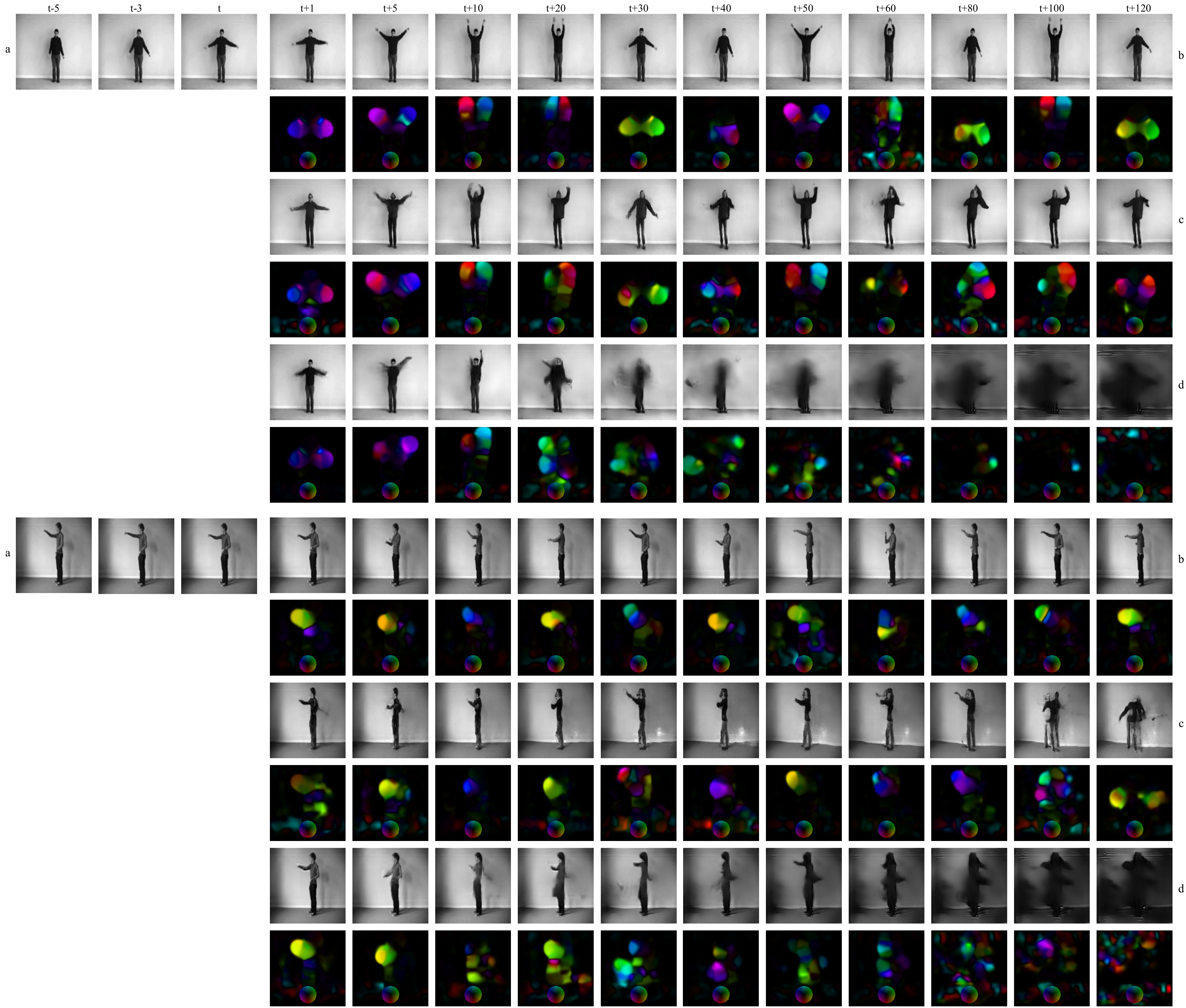}
	\caption{Prediction results of the 120 step long-term predictions for all three datasets. a: Input, b: Ground Truth, c: FutureGAN (ours), d: MCNet \citep{Villegas2017}.}
	\label{kthaction_deeppred120}
\end{figure}

\begin{figure}[h]
	\centering
	\subfigure{
		\includegraphics[width=0.3\linewidth]{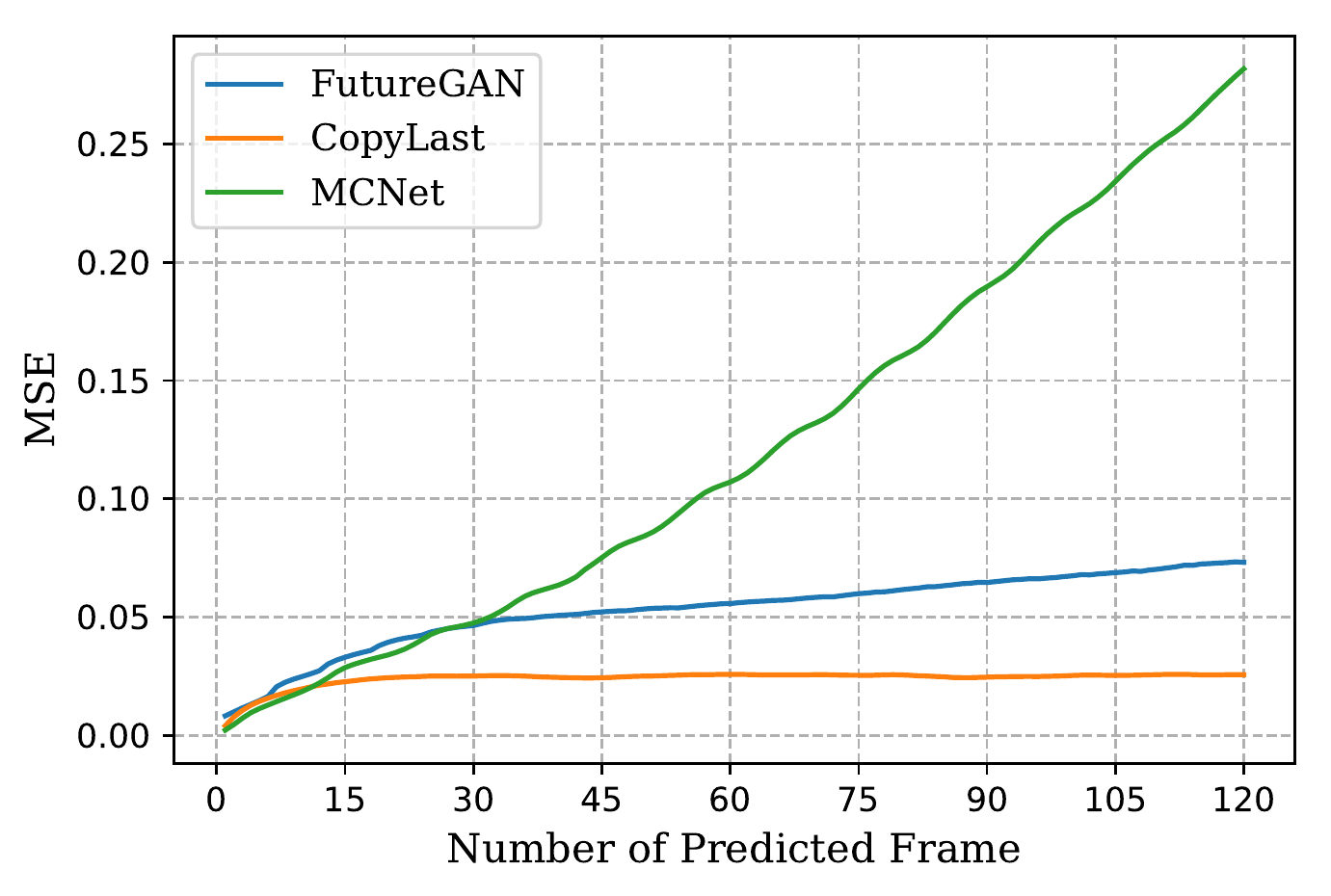}
		\label{kthaction_deeppred120_mse}}
	\setcounter{subfigure}{0}
	\subfigure{
		\includegraphics[width=0.3\linewidth]{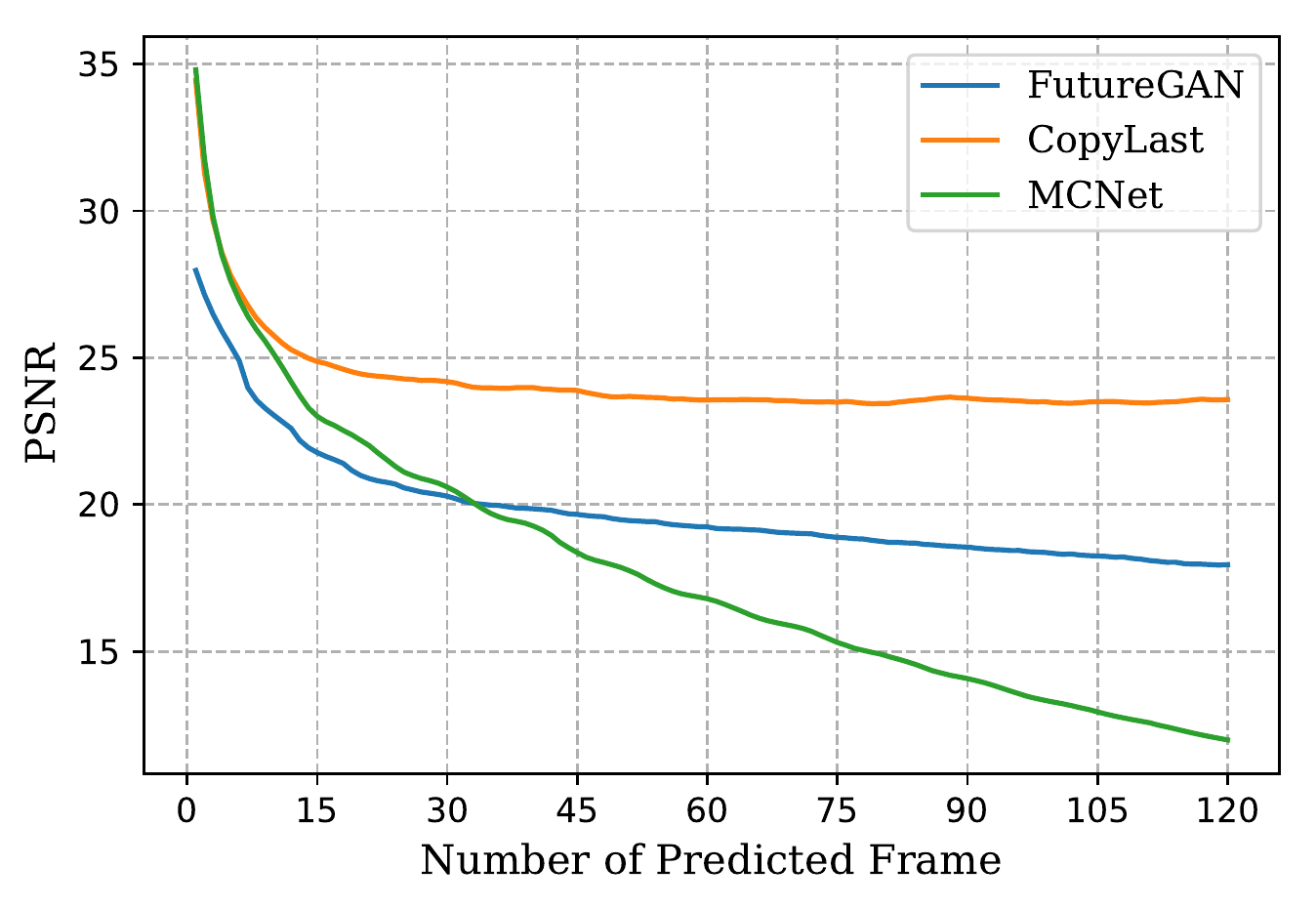}
		\label{kthaction_deeppred120_psnr}}
	\subfigure{
		\includegraphics[width=0.3\linewidth]{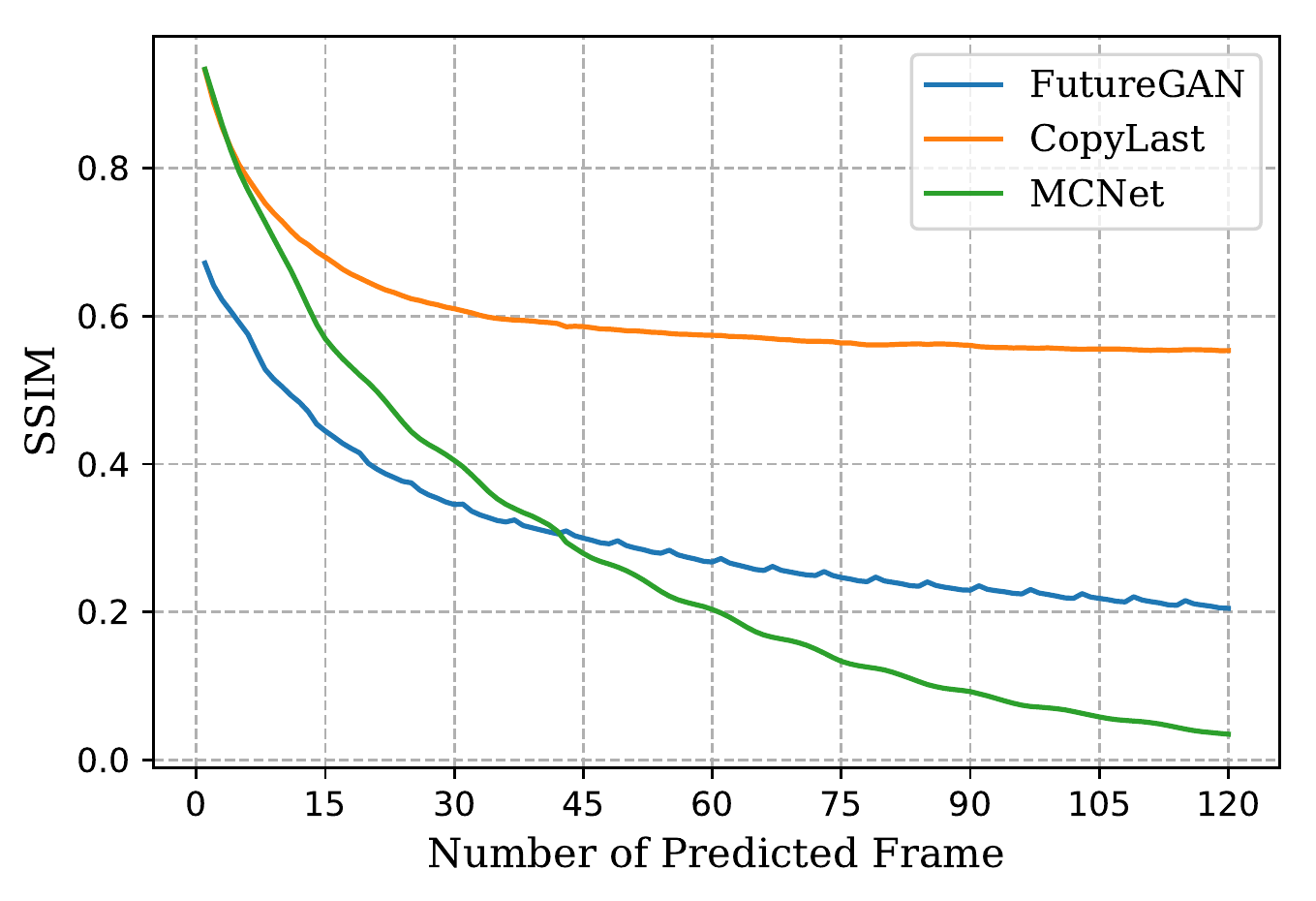}
		\label{kthaction_deeppred120_ssim}}		
	\caption{Quantitative results per predicted frame for the full KTH Action test split.}
	\label{kthaction_deeppred120_quantitative}
\end{figure}

\newpage
\subsection{Cityscapes} \label{appendix-cityscapes}
\begin{figure}[h]
	\centering
	\includegraphics[width=0.95\linewidth]{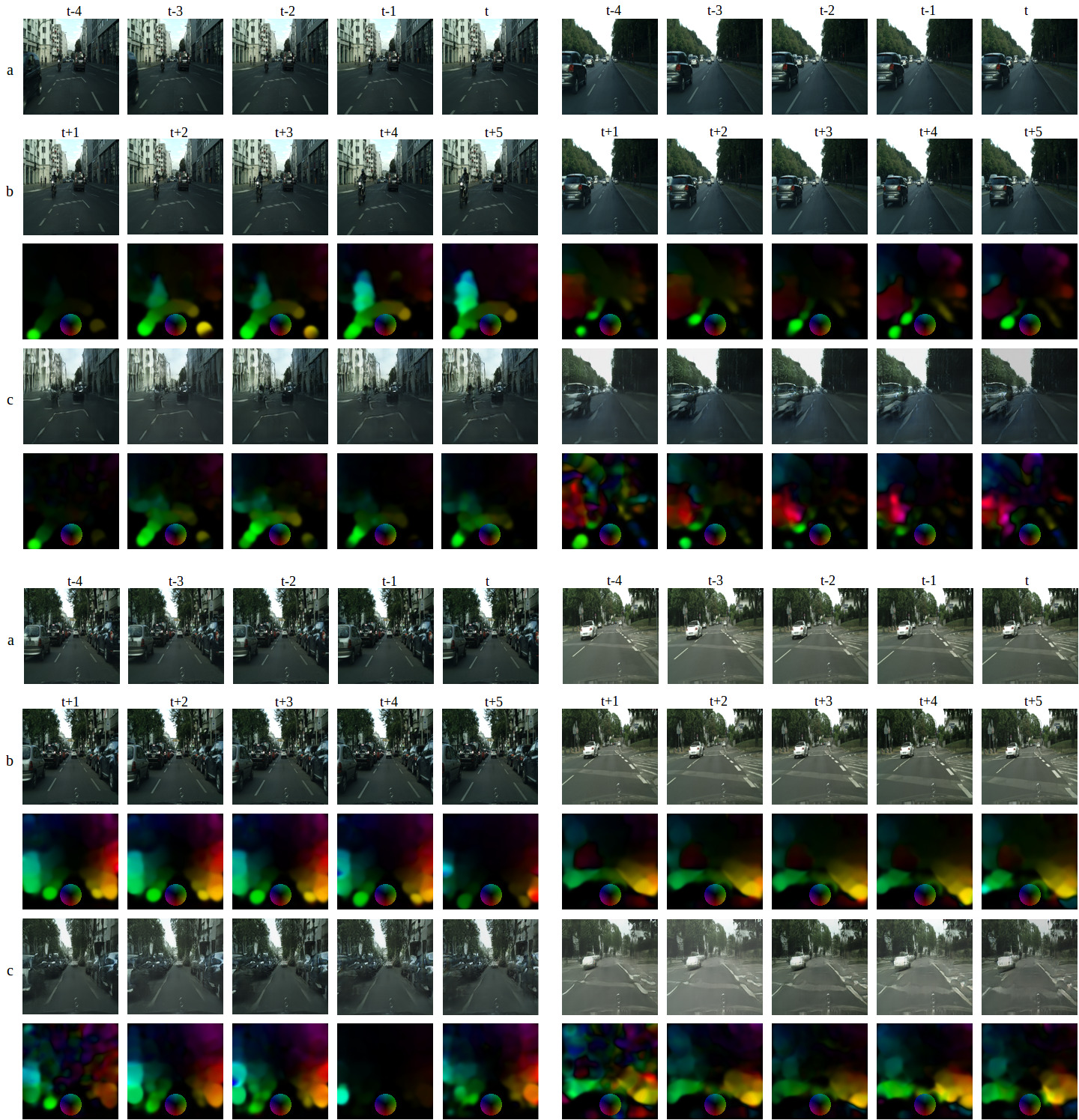}
	\caption{Prediction results for the Cityscapes test sequences. a: Input, b: Ground Truth, c: FutureGAN (ours)}
	\label{cityscapes_opticalflow}
\end{figure}

\begin{figure}[h]
	\centering
	\subfigure{
		\includegraphics[width=0.3\linewidth]{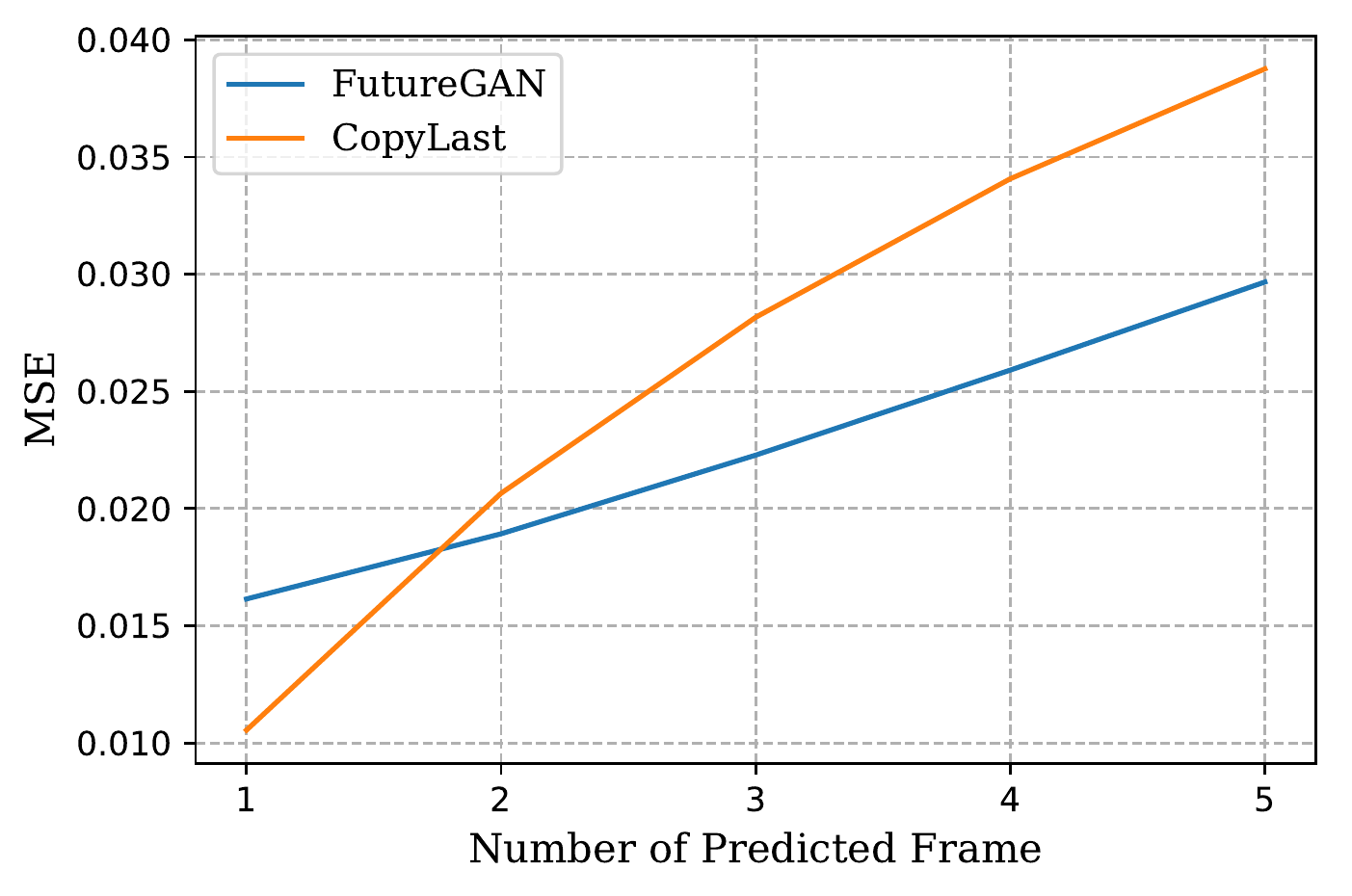}
		\label{cityscapes_mse}}
	\setcounter{subfigure}{0}
	\subfigure{
		\includegraphics[width=0.3\linewidth]{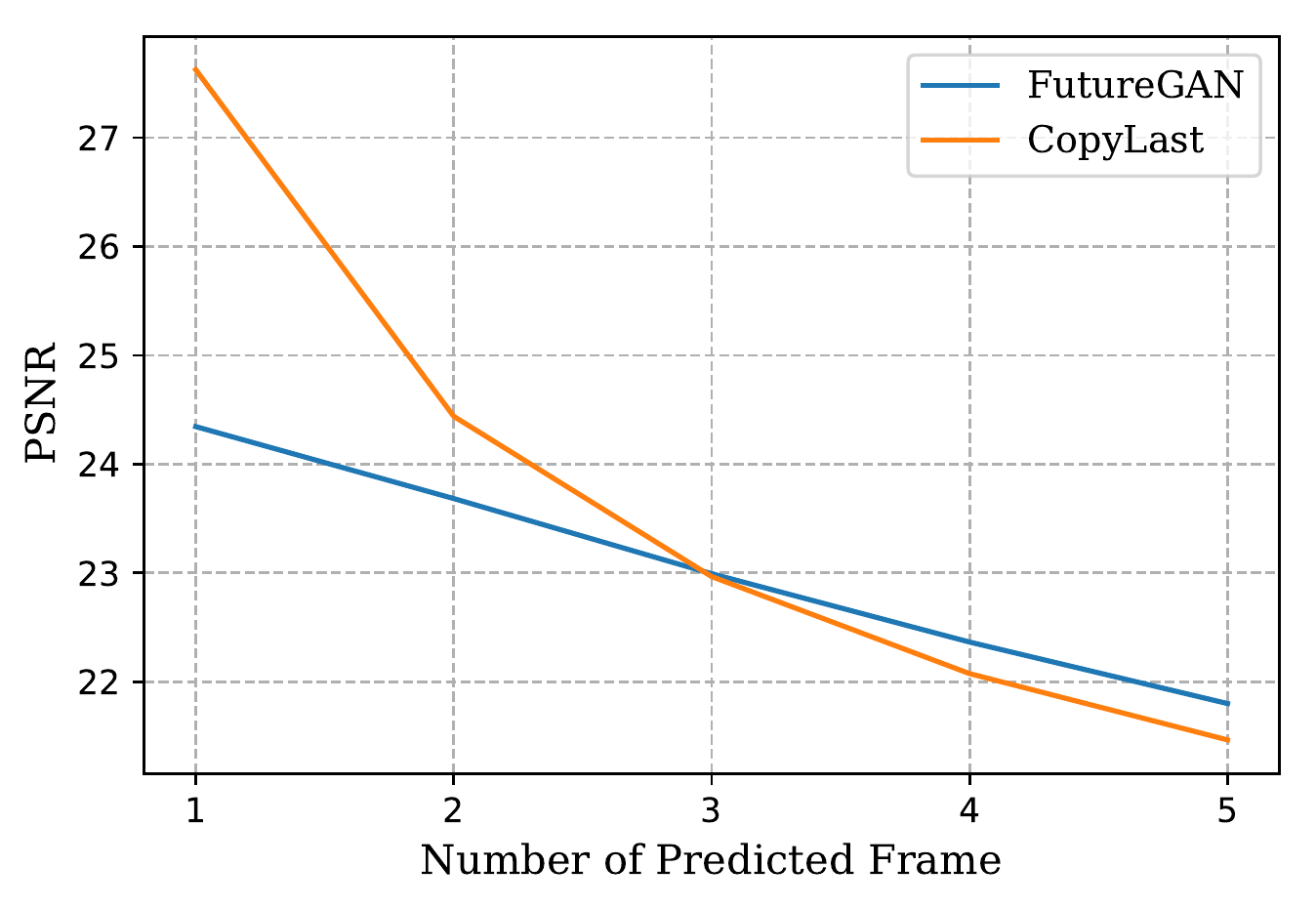}
		\label{cityscapes_psnr}}
	\subfigure{
		\includegraphics[width=0.3\linewidth]{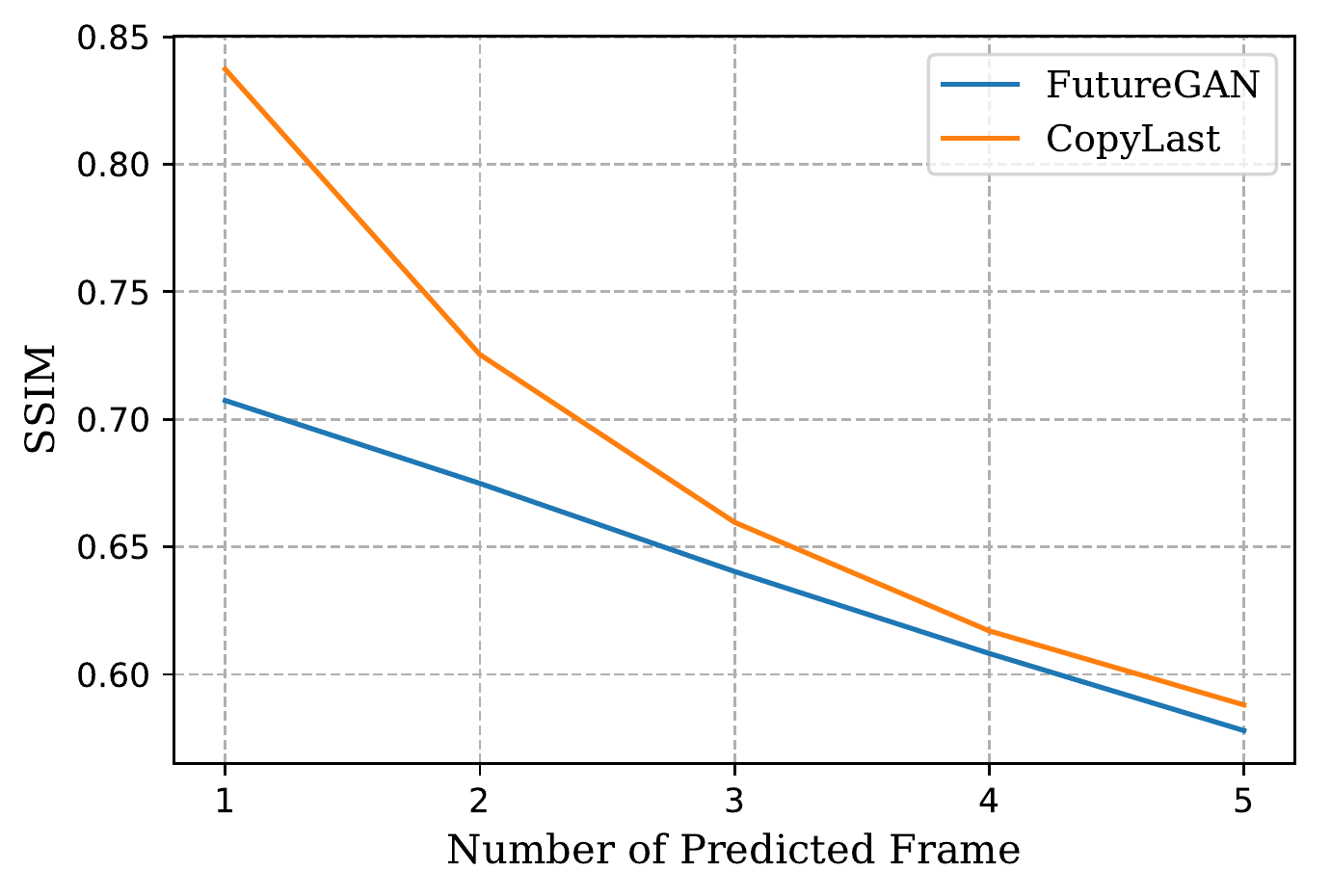}
		\label{cityscapes_ssim}}		
	\caption{Quantitative results per predicted frame for the full Cityscapes test split.}
	\label{cityscapes_quantitative}
\end{figure}

\newpage
\section{Discriminator Structure} \label{appendix-dsicriminator}
\begin{figure}[h]
	\centering
	\includegraphics[width=\linewidth]{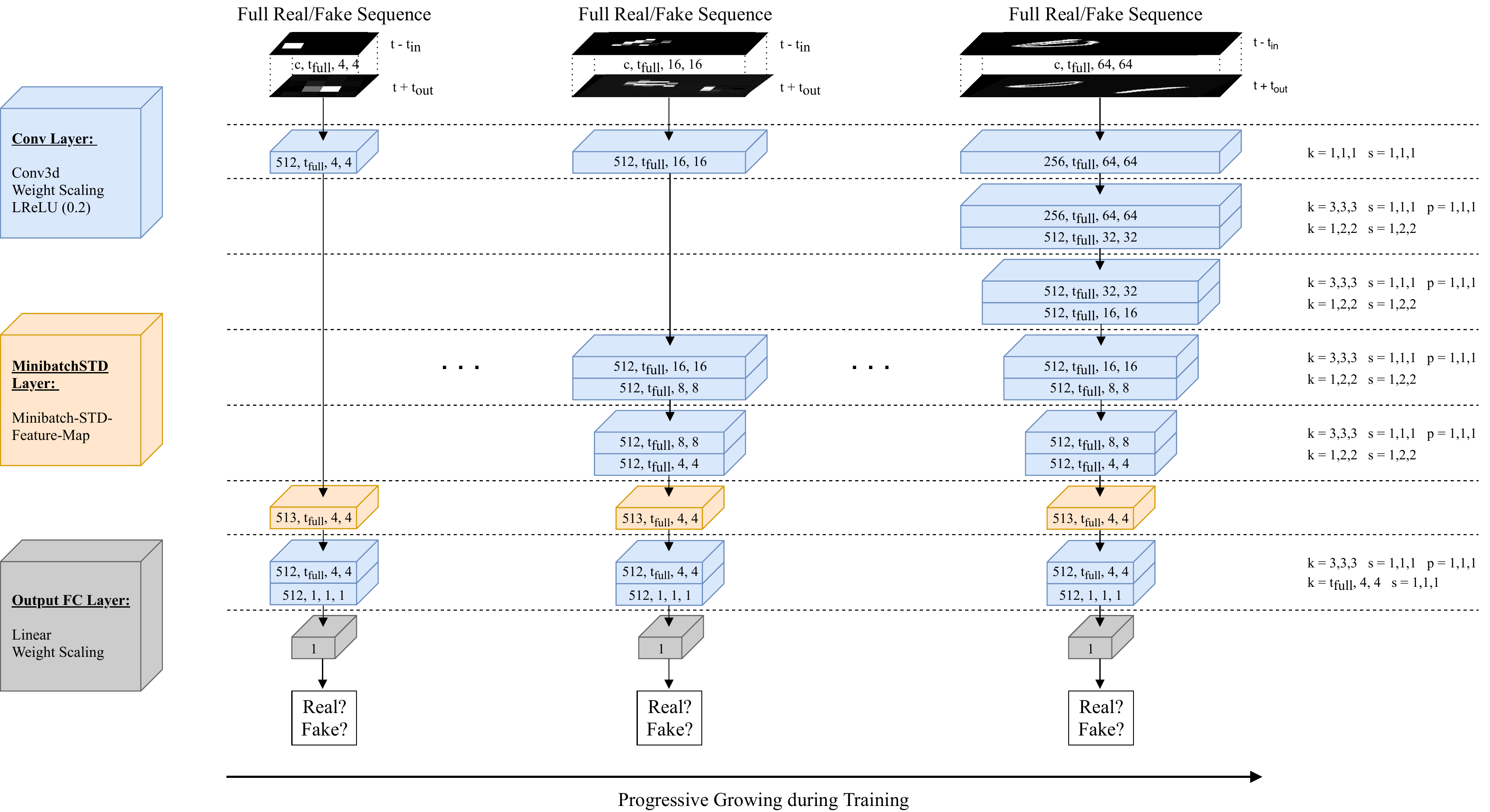}
	\caption{\textbf{FutureGAN discriminator during training.} 
		We initialize our model to take a set of \SI{4x4}{\px} resolution frames and output frames of the same resolution.
		During training, layers are added progressively to double the resolution after a specified number of iterations. 
		The resolution of the input frames always matches the resolution of the current state of the network.
		This figure illustrates the growth progress of the discriminator exemplary for the MovingMNIST dataset with a final resolution of \SI{64x64}{\px}. 
		Intermediate \SI{8x8}{\px} and \SI{32x32}{\px} resolution steps are left out for visual clarity.}
	\label{discriminator}
\end{figure}

\end{document}